\documentclass[11pt]{article}
\usepackage[final]{acl}
\usepackage{times}
\usepackage{latexsym}
\usepackage[T1]{fontenc}
\usepackage[utf8]{inputenc}
\usepackage{microtype}
\usepackage{inconsolata}
\usepackage{graphicx}
\usepackage{booktabs}
\usepackage{amsmath, amssymb}
\usepackage{fancyvrb}
\usepackage{fvextra}
\usepackage{multirow}
\usepackage[most]{tcolorbox}

\usepackage[ruled, vlined, linesnumbered]{algorithm2e}
\usepackage{comment}
\newcommand{\system}{\textsc{TripPulse}}
\title{\system{}: Multi-Agent Travel Planning with Review-Grounded Reasoning}

\author{Priyanshu Karmakar$^{1\dagger}$,
Borru Vijay Sai$^{1\dagger}$, Shubhojit Mallick$^2$,  Abhik Jana$^1$, Shreya Ghosh$^1$,\\\textbf{Manish Gupta$^2$} \\
  $^1$IIT Bhubaneswar, India\quad $^2$Microsoft, India\\
\texttt{{a24cs08008,22CS01076,abhikjana,shreya}@iitbbs.ac.in}\\
  \texttt{{shubhojit.mallick,gmanish}@microsoft.com} 
}

\begin{document}
\maketitle
\renewcommand{\thefootnote}{$\dagger$}
\footnotetext{The first two authors made equal contributions.}
\renewcommand{\thefootnote}{\arabic{footnote}}
\begin{abstract}

Travel itinerary generation requires balancing strict spatio-temporal constraints with human preferences. Existing LLM-based planners mainly rely on structured attributes and predefined traveler personas, but real travel decisions are often shaped by reviews that reveal experiential factors such as comfort, safety, service quality, ambiance, crowding, and hidden risks absent from structured databases. Incorporating such review information is therefore critical to realistic, user-centric itinerary generation. We propose \system{}\footnote{Captures experiential signals from reviews beyond rigid hard constraints.}, a multi-agent framework for review-grounded travel planning. Instead of relying on a monolithic planner (and face context and reasoning bottlenecks), \system{}\footnote{Code Base: \url{https://github.com/VijaySaiBorru/TripPulse}} decomposes itinerary generation into specialized agents (each operating over localized contexts) for accommodations, transportation, meals, attractions, and events, coordinated through a global orchestrator with scheduling mechanisms that enforce temporal and budget feasibility. We augment \textsc{TripCraft} with 100K+ real-world reviews and introduce Review-Grounded Persona Alignment (RGPA), an LLM-as-a-Judge metric for evaluating alignment with human-centric travel experiences. Experiments across multiple trip durations and diverse proprietary and open-source models show that \system{} maintains strong constraint satisfaction while generating more personalized and experientially grounded itineraries.

\end{abstract}

\section{Introduction}

Travel itinerary generation is challenging, requiring the joint satisfaction of strict spatio-temporal constraints and nuanced user preferences. Plans must coordinate transportation, hotels, and activities under budget and scheduling constraints while also optimizing subjective factors like safety, comfort, atmosphere and suitability, making it a complex real-world reasoning task.

Large Language Models (LLMs) show strong planning abilities~\citep{wei2022chain,yao2023react}, and benchmarks such as TravelPlanner~\citep{xie2024travelplanner}, TripCraft~\citep{chaudhuri2025tripcraft} and TripTide~\citep{karmakar-etal-2026-triptide} evaluate them under fine-grained constraints and diverse traveler profiles. However, most existing methods rely on \emph{monolithic prompting}, forcing a single model to handle multiple sub-tasks (entity selection, preference reasoning, scheduling, and constraint satisfaction), which often leads to hallucinations and violations of temporal or spatial feasibility as complexity grows. Recent agentic approaches~\citep{choi2026atlas} 
address reasoning overload via task decomposition, yet miss the experiential signals 
central to real travel decisions.

Another key limitation is \textit{how preferences are modeled}. Most approaches rely on structured attributes, while real travel decisions depend on experiential signals from reviews, such as safety, crowding, and contextual suitability. However, integrating reviews is challenging because they are large and noisy, and directly adding them to monolithic LLMs increases context load and reduces planning reliability.

\begin{figure*}[]
\centering
\includegraphics[width=1.1\textwidth]{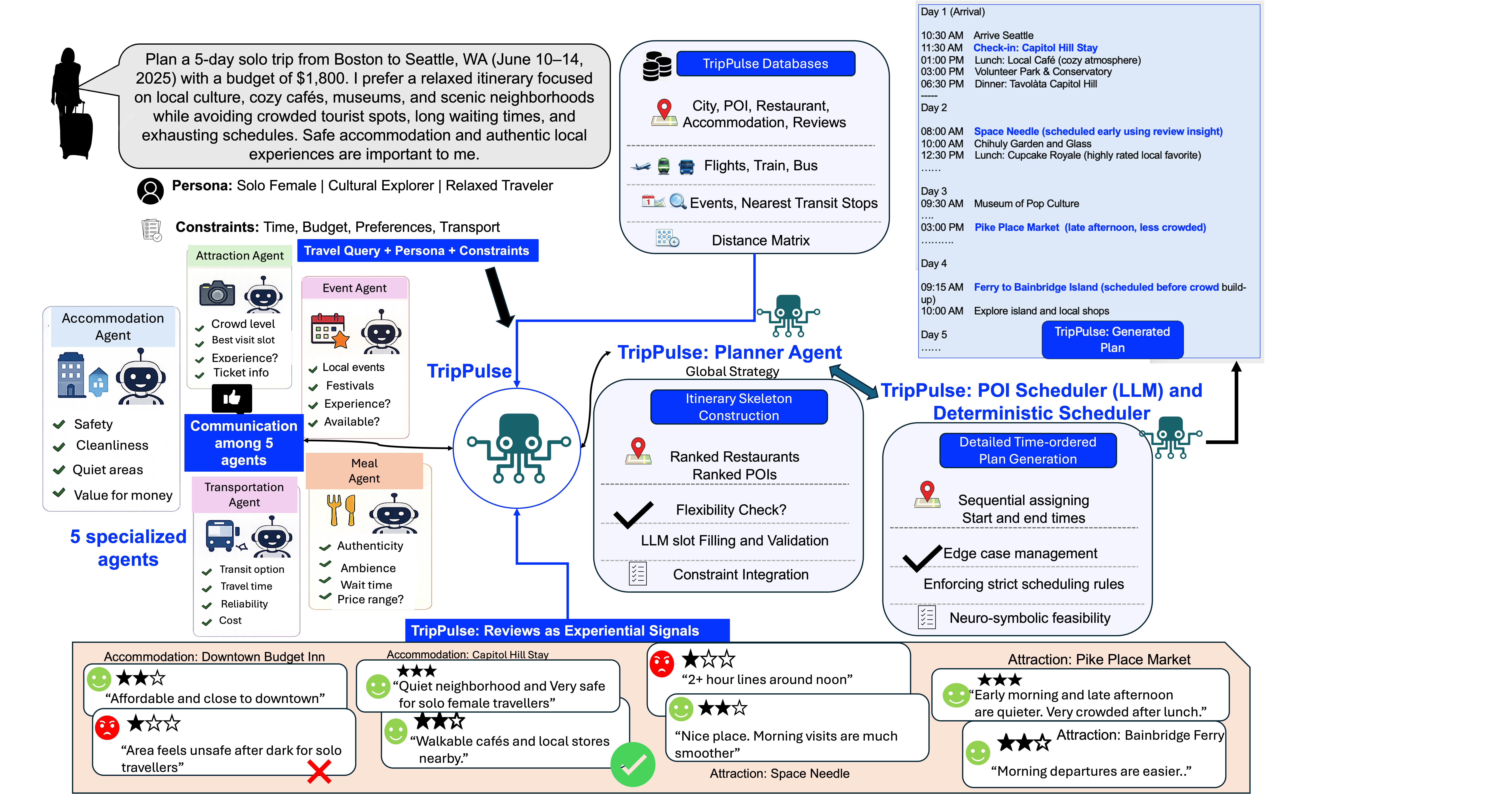}
\caption{The \system{} architecture. Planning is decomposed into 5 domain-specific agents, operating on localized contexts, orchestrated by a global orchestrator. The final itinerary is constructed via either an LLM-based POI Scheduler or a deterministic algorithmic scheduler to guarantee constraint satisfaction.}
\label{fig:framework}
\end{figure*}

We argue that leveraging review information requires restructuring the planning process. We propose \system{}, a multi-agent framework for \emph{review-grounded travel planning}, where domain-specific agents (accommodations, transportation, meals, attractions, and events) operate on localized contexts and are coordinated by a global orchestrator, while a downstream scheduling module enforces strict temporal and budget feasibility. Beyond traditional constraint satisfaction metrics, we introduce Review-Grounded Persona Alignment (RGPA), an LLM-as-a-Judge metric for evaluating alignment with experiential signals from reviews.

Our key contributions are as follows:
(1) \textbf{Review-augmented travel benchmark.}
We extend TripCraft with 100K+ Airbnb/TripAdvisor reviews, distilled into concise \textit{pros/cons} to capture experiential signals.

(2) \textbf{Review-grounded multi-agent framework}: \system{}, a modular architecture with context-isolated agents that integrates reviews while ensuring temporal and budget feasibility.
(3) \textbf{Human-centric evaluation}: \textit{RGPA}, an LLM-as-a-Judge metric for personalization, experiential quality, risk avoidance, and overall traveler satisfaction. We evaluate \system{} across 3/5/7-day trips with GPT-5~\citep{singh2026openaigpt5card}, Llama 3.1~\citep{grattafiori2024llama3herdmodels}, Phi-4-mini-instruct~\citep{microsoft2025phi4minitechnicalreportcompact}, Mistral-Nemo~\citep{mistral2024nemo}, Qwen-2.5-7B-instruct~\citep{qwen2025qwen25technicalreport}, and DeepSeek-R1~\citep{Guo_2025}.

\section{Related Work}
\noindent\textbf{LLM-Based Planning and Agents}: LLMs show strong reasoning capabilities for structured decision-making \citep{wei2022chain, yao2023react}, improved through chain-of-thought \citep{wei2022chain}, Tree-of-Thought \citep{yao2023tree}, and self-reflection \citep{shinn2024reflexion}. Multi-agent frameworks such as MetaGPT \citep{hong2024metagpt} and CAMEL \citep{li2023camel} further decompose tasks for efficiency. However, LLMs still struggle with long-horizon constraint satisfaction \citep{valmeekam2023large}, and scaling methods such as PDDL (Planning Domain Definition Language) grounding for preference-driven travel remain challenging \citep{liu2023llmpddl}. 

\noindent\textbf{Travel Planning Benchmarks and Systems}: Benchmarks like  TravelPlanner \citep{xie2024travelplanner} and TripCraft \citep{chaudhuri2025tripcraft} frame itinerary generation as a fine-grained multi-constraint problem \citep{xie2024travelplanner, chaudhuri2025tripcraft}.Recent agentic approaches~\citep{choi2026atlas} attempt to address reasoning overload through task decomposition, yet remain limited to structured attributes with no mechanism for review-grounded entity selection Retrieval-based methods such as TP-RAG \citep{ni2025tprag} improve point-of-interest (POI) selection but rely on structured data or past trajectories, and overlook rich review signals. Existing evaluations focus on constraint satisfaction and routing metrics \citep{chaudhuri2025tripcraft, xie2024travelplanner}, but fail to capture experiential quality. The LLM-as-a-Judge paradigm provides a solution for open-ended evaluation \citep{zheng2023judging}; we extend it with review-grounded signals to better measure human-centric quality. 

\noindent\textbf{Hybrid and Formal Planning Approaches}: Hybrid approaches combine LLMs with symbolic solvers such as Satisfiability Modulo Theories (SMT) to improve feasibility \citep{BT18, hao2025formalplanning}. While effective for constraints, they miss semantic nuances such as comfort and safety without extensive manual ontologies. Our approach combines agentic reasoning with a deterministic backend to address this gap. 

\noindent\textbf{Review-Aware Recommendation Systems}: Review-driven recommendation techniques highlight experiential attributes using structured summaries \citep{wei2025learning}, complementing traditional POI methods that mitigate data sparsity \citep{10.1145/3178876.3186070}. We are the first to port these distilled review summaries structured as \textit{Pros} and \textit{Cons} into constraint-aware planning.

\noindent\textbf{Our Approach}: \system{} uses agentic decomposition for review-grounded planning. By assigning specialized agents to localized contexts, distilled review signals can be processed efficiently without overwhelming a single model. This combines LLM flexibility with deterministic constraint enforcement to unify structural feasibility and user preference modeling. Detailed discussions are in App.~\ref{app:relatedWork}.

\section{\system{}}

In this section, we present \system{}, a multi-agent framework for spatio-temporal travel planning. Itinerary generation is decomposed across specialized agents operating on localized contexts. Their outputs are coordinated by an orchestration module and passed to a final planning backend.

To study the trade-off between generative flexibility and symbolic accuracy, we compare two separate planning pipeline configurations: (1) an LLM-based planner-scheduler pipeline and (2) a deterministic scheduler enforcing strict temporal feasibility. These are evaluated independently, without dynamic routing during inference, to analyze how varying levels of LLM autonomy impact itinerary reliability.

\subsection{Problem Formulation}

A travel query is defined as $q = (c_s, C_d, W, k, B, C_{local}, \pi)$, where $c_s$ is the source city, $C_d$ the destination set, $W$ the travel window, $k$ the number of travelers, $B$ the global budget constraint, $\pi$ the traveler persona, and $C_{local}$ the query-specific local constraints (e.g., room types, smoking rules, transport exclusions) used in TripCraft and TravelPlanner.

Let $\mathcal{E}$ denote candidate travel entities such as accommodations, restaurants, attractions, transport, and events. The goal is to generate a time-ordered itinerary $I = {(e_i, t_i^{start}, t_i^{end})}_{i=1}^{N}$, where each $e_i \in \mathcal{E}$ is a selected entity with scheduled start and end times. Itineraery contains $N$ entities.

A valid itinerary must satisfy: (1) Budget Constraint: total cost for $k$ travelers $\leq B$; (2) Temporal Feasibility: activities must lie within $W$, follow chronological order, and respect transit buffer $\Delta_{transit}$ (Table \ref{tab:temporal_constants} in App.~\ref{app:hyperparams}); (3) Local Constraint Satisfaction: no violations of attribute rules in $C_{local}$; and (4) Preference Alignment: maximize review-based quality for persona $\pi$.

Our objective is to maximize review-grounded quality while strictly satisfying all structural (valid database entities, fixed dates and cities), temporal, and local constraints.

\subsection{Overview of \system{}}

Fig.~\ref{fig:framework} shows the architecture of \system{}, and Algorithm~\ref{alg:agentic_tripcraft} details its execution flow. Given a user query with preferences, duration, and budget, the system generates time-ordered itineraries through a multi-stage agentic pipeline consisting of: (1) five domain-specific reasoning agents, (2) a \texttt{Global Orchestrator}, and (3) a scheduling backend.

Each agent processes relevant subsets of the travel database and proposes candidate entities. The \texttt{Global Orchestrator} coordinates agents sequentially (Phases 1-4 of Algorithm~\ref{alg:agentic_tripcraft}), manages budget allocation, and enforces global constraints such as duration and total budget. The collected entity selections are converted into an intermediate representation and passed to the scheduling backend (Phase 5), which builds the final itinerary using either an LLM-based scheduler (\texttt{FinalScheduleAgent}) or a deterministic (algorithm) scheduler (\texttt{FinalScheduleBuilder}). This decomposition reduces reasoning complexity and context size, improving reliability and grounding while lowering the risk of hallucinations.

\begin{algorithm}[t]
\scriptsize
\DontPrintSemicolon
\KwIn{Natural language travel query $q$, Travel database $\mathcal{DB}$, Temporal rules $\mathcal{R}_{temp}$ (Table~\ref{tab:temporal_constants})} 
\KwOut{Feasible Spatio-Temporal Itinerary $I$}
\textbf{Intermediate Variables:} $A_{acc}$ (Selected Accommodation), $A_{trans}$ (Selected Transport), $R_{rank}$ (Ranked Restaurants), $T_{rank}$ (Ranked Attractions), $E_{opt}$ (Optional Events), $S$ (Trip Skeleton), $I_{unsched}$ (Unscheduled Itinerary).\;
\textbf{Phase 1: Query Unpacking} \;
Extract attributes $c_s, C_d, W, k, B, C_{local}, \pi$ from $q$ \;
\textbf{Phase 2: Accommodation Selection} \;
\For{each destination city $c \in C_d$}{
    $A_{acc}[c] \leftarrow \text{AccoAgent}(C_{local}, \pi, \mathcal{DB}_{acc}[c])$ \;
}
\textbf{Phase 3: Global Transportation Planning} \;
$A_{trans} \leftarrow \text{TransportAgent}(C_{local}, \pi, \mathcal{DB}_{trans}, W, B)$  \;
\textbf{Phase 4: City-Level Domain Agents} \;
\For{each destination city $c \in C_d$}{
    $R_{rank}[c] \leftarrow \text{MealsAgent}(C_{local}, \pi, \mathcal{DB}_{rest}[c])$ \;
    $T_{rank}[c] \leftarrow \text{AttractionAgent}(C_{local}, \pi, \mathcal{DB}_{attr}[c])$ \;
    $E_{opt}[c] \leftarrow \text{EventsAgent}(C_{local}, \pi, \mathcal{DB}_{event}[c])$ \;
}
\textbf{Phase 5: Skeleton Construction \& Scheduling} \;
$S \leftarrow \text{GenerateSkeleton}(A_{acc}, A_{trans}, W)$ \;
\If{Scheduling Pathway == LLM-based}{
    $I_{unsched} \leftarrow \text{FillSkeleton}(S, R_{rank}, T_{rank}, E_{opt})$ \;
    $I \leftarrow \text{LLM-POIScheduler}(I_{unsched}, \mathcal{R}_{temp})$ \;
    }
\Else{
    $I \leftarrow \text{AlgoScheduler}(S, R_{rank}, T_{rank}, E_{opt}, \mathcal{R}_{temp})$ \;
}
\textbf{return} $I$ \;
\caption{\system{} Itinerary Generation Pipeline (Notations in  App.~\ref{app:notations})}
\label{alg:agentic_tripcraft}
\end{algorithm} 

\subsection{Dataset Construction}
\label{sec:dataset_construction}

\noindent\textbf{Gathering Reviews.}
To extend TripCraft from a structural to an experiential benchmark, we augmented it with review data collected between Dec 2025 and Jan 2026 across 140 U.S. cities. Accommodation reviews were scraped from Airbnb\footnote{\url{https://www.airbnb.com/}}, while restaurant and attraction reviews came from TripAdvisor\footnote{\url{https://www.tripadvisor.com/}}. Crawlers targeted the exact entity IDs in TripCraft, ensuring a strict 1:1 mapping between database entities and reviews. All reviews were publicly available and anonymized to remove Personally Identifiable Information (PII).

\noindent\textbf{Extracting Pros/Cons from Reviews.} We leveraged \texttt{Qwen/Qwen3-4B-Instruct} to convert unstructured reviews into structured experiential signals. For each entity, up to five reviews were aggregated into a single context window. A zero-shot prompt was used to summarize reviews into advantages and disadvantages using a strict JSON schema: \texttt{{``Pros'': [], ``Cons'': []}} using greedy decoding with \texttt{max\_new\_tokens=256}. We implemented a parsing layer to correct formatting deviations by removing markdown, normalizing key names, and converting singular strings into arrays. Failed generations were replaced with empty feature lists to ensure stable large-scale processing. The resulting ``Pros'' and ``Cons'' were merged back into the travel database. Table~\ref{tab:dataset_stats} details the scale of our augmented review dataset following this processing pipeline.

\begin{table}[h]
\centering

\scriptsize
\begin{tabular}{llrr}
\toprule
\textbf{Domain} & \textbf{Source} & \textbf{Entities} & \textbf{Reviews} \\
\midrule
Accommodations & Airbnb & 2,395 & 18,094 \\
Restaurants & TripAdvisor & 3,765 & 42,541 \\
Attractions & TripAdvisor & 4,586 & 50,993 \\
\midrule
\textbf{Total} & & \textbf{10,746} & \textbf{111,628} \\
\bottomrule
\end{tabular}%
% }
\caption{Statistics of the scraped and processed review dataset used to augment the TripCraft benchmark.}
\label{tab:dataset_stats}
\end{table}
\begin{comment}

\begin{figure}[h]
\centering
\includegraphics[width=\linewidth]{review_pipeline.png}
\caption{Our multi-model review processing pipeline. Unstructured reviews are processed via RoBERTa (sentiment), generative LLMs (summarization), and MPNet (semantic alignment) to generate structured Quality ($Q$), Risk ($R$), and Persona ($A$) signals.}
\label{fig:review_pipeline}
\end{figure}
\end{comment}

\subsection{Global Orchestrator}

The Global Orchestrator coordinates domain-specific agents while enforcing global constraints such as budget, duration, and city transitions.
The orchestrator follows a hybrid sequential-parallel design. Accommodation, Transportation, and Meals agents execute sequentially because they share the global budget constraint ($B$). They draw from a shared, monotonically decreasing budget pool. In contrast, Attraction and Event agents act as preference-ranking modules and are executed asynchronously in parallel since they do not modify the budget state.

To ensure financial feasibility, the orchestrator allocates the budget progressively rather than exposing the full amount to all agents. After accommodation selection, the remaining budget is recalculated, with 85\% assigned as the upper bound for transportation and the remaining 15\% reserved for meals. This deterministic allocation guarantees budget compliance without expensive recursive retries. Retries are used only for parsing or API failures, and surplus budget can optionally enable accommodation upgrades.

The orchestrator aggregates agent outputs into a structured trip plan and passes it to the final scheduling stage. By centralizing constraint management with deterministic state updates, it enables modular reasoning while preserving global feasibility.

\subsection{Domain-Specific Agents}

The first stage of the pipeline consists of five domain-specific agents that reason over the query $q$. Except for the Transportation Agent, which operates globally across the itinerary ($I$), all agents function at the destination-city level ($c \in C_d$) using localized data.

\noindent\textbf{Accommodation Agent.} The Accommodation Agent selects lodging ($A_{acc}[c]$) for each city using traveler persona ($\pi$), local constraints ($C_{local}$), and accommodation data ($\mathcal{DB}_{acc}$). It combines structural attributes (pricing, room type, occupancy limits, and house rules) with review-derived ``Pros'' and ``Cons''. By synthesizing these qualitative advantages and disadvantages with the structural constraints, the agent chooses a feasible, affordable and preference-aligned property.

\noindent\textbf{Transportation Agent.} This agent constructs a globally consistent routing strategy ($A_{trans}$) across cities using transportation data ($\mathcal{DB}_{trans}$), including flights, taxis, and self-driving routes. Unlike the city-level agents that rank multiple candidates, this agent locks in a single, globally cohesive transportation strategy to ensure temporal consistency across all transit legs while satisfying the global budget ($B$).

\noindent\textbf{Meals Agent.} For each city, the Meals Agent retrieves restaurant candidates ($\mathcal{DB}{rest}[c]$) constrained by budget, and augments them with review-based ``Pros'' and ``Cons''. Using cuisine constraints and persona preferences ($\pi$), it generates a ranked list of restaurants ($R{rank}[c]$).

\noindent\textbf{Attraction Agent.} The Attraction Agent ranks candidate attractions ($\mathcal{DB}_{attr}[c]$) to obtain a final ranked list ($T_{rank}[c]$) by evaluating experiential review features such as cultural value, scenery, overcrowding or safety concerns, against the traveler persona ($\pi$).

\noindent\textbf{Events Agent.} The Events Agent filters events ($\mathcal{DB}_{event}[c]$) within the travel window ($W$) and selects optional activities ($E_{opt}[c]$) aligned with user preferences. To reduce scheduling conflicts, it limits selection to at most one event per day.

\subsection{LLM-Based Scheduling Pathway}

The LLM-based pipeline generates the final itinerary ($I$) in two stages: first, constructing a populated itinerary skeleton ($I_{unsched}$), and second, converting it into a detailed, time-ordered schedule via a POI Scheduler.

\noindent\textbf{Skeleton Generation and Slot Filling}. 
The system first builds a rigid day-level skeleton ($S$) using selected accommodations ($A_{acc}$), transportation schedules ($A_{trans}$), and time-bound events ($E_{opt}$). Transportation timings define the available activity slots for each day. To ensure convergence, transportation decisions are treated as immutable, preventing cyclic dependencies where modifying flights would invalidate downstream budget allocations and ranked restaurant candidates.

Once the skeleton is fixed, an LLM fills the remaining slots by selecting restaurants and attractions from ranked candidate lists ($R_{rank}, T_{rank}$). The plan is then validated against the TripCraft database to detect hallucinations, duplicates, infeasible ordering, and transportation conflicts.

Invalid components are selectively regenerated by the LLM, producing a validated unscheduled itinerary ($I_{unsched}$).

\noindent\textbf{POI Scheduler.} This converts $I_{unsched}$ into a time-ordered itinerary by assigning explicit start ($t_i^{start}$) and end times ($t_i^{end}$) to activities. This is done using state-aware prompts containing the selected entities and execution rules ($\mathcal{R}_{temp}$) governing activity ordering and timing. The LLM enforces temporal feasibility between activities, transit buffers ($\Delta_{transit}$), max daily attraction limits, and meal ordering.

This results in a hybrid architecture in which LLMs handle semantic planning for selecting and scheduling activities, while deterministic validation ensures temporal consistency and mathematical feasibility.

\subsection{Deterministic Scheduling Pathway}

As an alternative to the LLM-based scheduling, our framework includes a rule-based Algorithmic Scheduler that constructs the final itinerary ($I$) entirely through programmatic execution.

\noindent\textbf{Bipartite Scheduling Logic.} 
Similar to the LLM-based pathway, the deterministic scheduler operates in two stages: skeleton construction and greedy population. It first creates the same day-level skeleton ($S$) by fixing accommodations ($A_{acc}$), transportation ($A_{trans}$), and time-bound events ($E_{opt}$), thereby defining available activity windows.

The scheduler then greedily fills remaining slots using the ranked restaurant and attraction lists ($R_{rank}$, $T_{rank}$). Instead of semantic reasoning, it iteratively inserts the highest-ranked feasible entities into open time windows.

\noindent\textbf{Constraint Enforcement.} During greedy insertion, the scheduler assigns explicit start ($t_i^{start}$) and end times ($t_i^{end}$) while enforcing temporal rules ($\mathcal{R}_{temp}$), including transit buffers ($\Delta{transit}$), meal timing constraints, and persona-aware duration scaling (e.g., extending the duration of an attraction visit for laidback travelers).

By replacing generative reasoning with deterministic execution, this baseline guarantees full constraint satisfaction while reducing scheduling latency and computational overhead.

\subsection{Agent Invocation Complexity}

While our multi-agent framework increases the number of LLM calls compared to monolithic prompting, it shifts the computational bottleneck through decomposition. Domain-specific agents operate independently on localized data, while global modules coordinate the overall itinerary.

This design keeps the \emph{context size per call bounded}, since each agent processes only a small domain-specific subset of the database (e.g., restaurants within one city) instead of the full travel catalog. As a result, the framework avoids large context windows and expensive attention costs, reducing hallucination risk and enabling smaller open-source models to achieve strong constraint satisfaction.

\section{Experimental Setup}

\subsection{Dataset and Models}

\noindent\textbf{Dataset}: We evaluate \system{} on TripCraft~\citep{chaudhuri2025tripcraft}, augmented with structured review-derived ``Pros'' and ``Cons'' (Section~\ref{sec:dataset_construction}). The benchmark provides structured travel queries that specify $C_d$, $W$, $B$, and $\pi$, along with deterministic databases of transportation schedules and hard constraint rules.

To study \system{}'s scalability and robustness, we generate itineraries for 3-day, 5-day, and 7-day trips. Longer horizons substantially increase scheduling complexity (due to several interconnected travel components), requiring the system to maintain strict temporal and budget feasibility while optimizing experiential quality from the augmented review signals.

\noindent\textbf{Models}: We evaluate \system{} across proprietary and open-weight LLMs across varying capability tiers, parameter scales, and cost regimes. These include proprietary GPT-5, large open-weight Llama-3.1-70B-Instruct, distilled reasoning model DeepSeek-R1-Distill-Qwen-14B, and efficient open-weight models (Phi-4, Mistral-Nemo-Instruct-2407, and Qwen-2.5-7B-Instruct). All models operate in a zero-shot setting without task-specific fine-tuning, ensuring the evaluation measures the effectiveness of the agentic framework rather than model-specific training.

\subsection{Evaluation Metrics}

We evaluate using 3 groups of metrics: constraint satisfaction metrics, temporal consistency metrics, and review-grounded experience metrics.

\noindent\textbf{Constraint Satisfaction Metrics.} We adopt these metrics from TripCraft: (1) Delivery Rate (Del)=\% of prompts yielding a completely formatted itinerary, (2) commonsense constraint satisfaction via commonsense pass rates: micro (CPR$_\mu$) and macro (CPR$_M$), (3) hard constraint satisfaction (e.g., budget limits, explicit accommodation rules) via hard constraint pass rate: micro (HCPR$\mu$) and macro (HCPR$_M$), and (4) Final Pass Rate (FPR), which measures itineraries satisfying all commonsense and hard constraints simultaneously.

\noindent\textbf{Temporal and Structural Metrics}. We also use TripCraft’s coherence metrics to evaluate realism: Temporal Meal Score ($T_m$) and Temporal Attraction Score ($T_a$)\footnote{We update this metric definition of TripCraft to make it more realistic as discussed in Appendix~\ref{app:temporal_metric}.} assess timing feasibility, while Ordering Score ($S_o$), Spatial Score ($S_s$), and Persona Alignment Score ($S_p$) measure logical activity sequencing, geographic routing and transit feasibility, and alignment with the traveler profile.

\noindent\textbf{Review-Grounded Evaluation Metric.}

To evaluate experiential quality beyond constraint satisfaction, we introduce the Review-Grounded Persona Alignment (RGPA) framework using an \textit{LLM-as-a-Judge} setup. The goal is to measure whether review integration produces itineraries that better match traveler preferences, are safer, and improve overall travel quality.

A judge model (GPT-5) compares itineraries generated with and without review signals. Given the same query, traveler persona, and structured review evidence, it evaluates itineraries across these dimensions: (1) Persona Alignment (on travel style, spending preference, and location interests), (2) Experiential Quality (using positive review signals such as atmosphere, service quality, uniqueness, and enjoyment), 

and (3) Overall Travel Satisfaction (considering both positive and negative experiential factors). The full prompt is provided in Appendix~\ref{app:llm_judge_prompt}. We also compute win rate as the \% of pairwise comparisons in which the review-grounded itinerary was preferred over the corresponding non-review itinerary by aggregating scores across the above 3 dimensions.

For itineraries $I_A$ and $I_B$, persona $\pi$, and review evidence $R$, the judge computes scores: $J(I_A, I_B, \pi, R) \rightarrow \{s_A^{(d)}, s_B^{(d)}\}_{d \in D}$,
where $D$ contains the evaluation dimensions and $s^{(d)} \in [1,10]$. Final scores for each dimension are averaged across all itinerary pairs. By incorporating review-derived experiential signals, RGPA enables a more holistic evaluation of personalized travel quality than traditional constraint-based metrics alone.

\begin{table}[t]
\centering
\scriptsize
\setlength{\tabcolsep}{1pt} 
% \resizebox{\textwidth}{!}{%
\begin{tabular}{lc cccccc cccccc}
\toprule
\multirow{2}{*}{} & \textbf{Duration} 
& \multicolumn{4}{c}{\textbf{Constraint Satisfaction (\%)}} 
& \multicolumn{6}{c}{\textbf{Temporal+Structural}} \\
\cmidrule(lr){3-6} \cmidrule(l){7-12}
& (day) & Del & CPR$_\mu$  & HCPR$_\mu$ & FPR & $T_m$ & $T_a$ & $\tilde{T}_a$ & $S_s$ & $S_p$ & $S_o$ \\
\midrule
\multicolumn{14}{l}{\textit{\textbf{Baseline: Monolithic Prompting (TripCraft~\cite{chaudhuri2025tripcraft})}}} \\
\midrule
\multirow{3}{*}{\rotatebox{90}{GPT5}} & 3& 100 & 81.54  & 94.49  & 1.45 & .79 & .17 & -- & .81 & .49 & .80 \\
               & 5& 100 & 67.47  & 92.92 & 0 & .84 & .22 & -- & .83 & .49 & .95 \\
               & 7& 99.70 & 57.28 & 90.12  & 0 & .86 & .22 & -- & .87 & .50 & .97 \\
\hline
\multirow{3}{*}{\rotatebox{90}{Phi4}} & 3& 92.60 & 47.69 & .0  & .0 & .24 & .22 & -- & .60 & .53 & .66 \\
      & 5& 99.56 & 43.86 & .0 & .0 & .53 & .13 & -- & .83 & .53 & .92 \\
      & 7& 97.79 & 37.22  & .0  & .0 & .51 & .14 & -- & .83 & .53 & .96 \\
\hline
\multirow{3}{*}{\rotatebox{90}{Qwen2.5}} & 3& 99.56 & 70.39  & 3.26  & .0 & .58 & .15 & -- & .869 & .51 & .79 \\
         & 5& 99.56 & 52.30  & .0  & .0 & .59 & .09 & -- & .742 & .51 & .92 \\
         & 7& 99.13 & 39.29 & .0  & .0 & .57 & .01 & -- & .734 & .52 & .96\\

\midrule
\multicolumn{14}{l}{\textit{\textbf{Baseline: Agentic Framework (ATLAS~\cite{choi2026atlas})}}} \\
\midrule
\multirow{3}{*}{\rotatebox{90}{Qwen2.5}} & 3& 100 & 53.22 & 1.27 & 0 & 0.28 & 0.07 & -- & 0.70 & 0.50 & 0.60 \\
         & 5& 100 & 37.1  & 0 & 0 & 0.19 & 0.03 & -- & 0.56 & 0.46 & 0.85 \\
         & 7& 100 & 34.14  & 0 & 0 & 0.20 & 0.05 & -- & 0.61 & 0.48 & 0.91\\         
\midrule
\multicolumn{14}{l}{\textit{\textbf{\system{} (Ours): LLM Scheduler}}} \\
\midrule
\multirow{3}{*}{\rotatebox{90}{GPT5}}& 3& 100 & 94.39  & \textbf{91.02} & 52.03 & \textbf{.87} & .13 & .36 & .86 & .51 & .80 \\
      & 5& 100 & 90.83  & 87.23 & 33.33 & .78 & .25 & .71 & .95 & .51 & .93 \\
      & 7& 100 & 80.69  & 87.72  & 5.42 & .74 & .25 & .73 & \textbf{\underline{.95}} & .50 & .96 \\
\hline
\multirow{3}{*}{\rotatebox{90}{Phi4}}& 3& 94.48 & 83.88 & 89.96  & 10.17 & .83 & .21 & .58 & .88 & \textbf{\underline{.52}} & .66 \\
      & 5& 89.20 & 75.74  & \textbf{88.86}  & 1.54 & .80 & .20 & .55 & .93 & .51 & .89 \\
      & 7& 92.77 & 71.10  & 86.88  & 1.20 & .80 & .18 & .53 & \textbf{\underline{.95}} & \textbf{\underline{.52 }} & .95 \\
\hline
\multirow{3}{*}{\rotatebox{90}{Qwen2.5}} & 3& 99.42 & 77.65  & 80.93 & .00 & .50 & .10 & .29 & .87 & \textbf{\underline{.52}} & .62 \\
         & 5& 96.91 & 67.25  & 82.74  & .00 & .46 & .15 & .43 & .95 & \textbf{\underline{.52 }}& .88 \vspace{1pt}\\
         & 7& 97.59 & 65.57 & 81.51  & .00 & .49 & .15 & .42 & .93 & \textbf{\underline{.52 }}& .94 \vspace{1pt}\\
\hline
\multirow{3}{*}{\rotatebox{90}{Llama3.1}}& 3& 95.63 & 77.06  & 82.90  & .29 & .62 & .15 & .42 & .89 & \textbf{\underline{.52}} & .67 \vspace{1pt}\\
         & 5& 85.19 & 62.65  & 72.34  & .0 & .59 & .12 & .33 & .93 & \textbf{\underline{.52}} & .89 \vspace{1pt}\\
         & 7& 99.70 & 69.67  & 83.25  & .0 & .56 & .11 & .31 & .92 & \textbf{\underline{.52}} & .95 \vspace{1pt}\\
\hline
\multirow{3}{*}{\rotatebox{90}{DS-R1}}& 3& 78.78 & 61.05  & 68.39  & 2.33 & .73 & .22 & .61 & \textbf{\underline{.90}} & \textbf{\underline{.52}} & .66 \\
         & 5& 53.40 & 35.50  & 45.15  & .0 & .72 & .20 & .58 & .93 & .51 & .88 \\
         & 7& 37.95 & 23.89  & 32.26  & .0 & .72 & .17 & .50 & .92 & .52 & .94 \\
\hline
\multirow{3}{*}{\rotatebox{90}{M-Nemo}} & 3& 98.84 & 77.06 & 85.61  & 1.16 & .56 & .15 & .43 & .89 & \textbf{\underline{.52}} & .65 \\
         & 5& 90.12 & 66.27  & 76.83 & .0 & .53 & .13 & .36 & .94 & \textbf{\underline{.52 }}& .89 \\
         & 7& 90.96 & 65.93  & 77.30  & .0 & .52 & .12 & .34 & .92 & \textbf{\underline{.52}} & .95 \\
\midrule
\multicolumn{14}{l}{\textit{\textbf{\system{} (Ours): Deterministic Algorithmic Scheduler}}} \\
\midrule
\multirow{3}{*}{\rotatebox{90}{GPT5}}& 3& 100 & 98.92  & 90.9  & \textbf{74.42} & .81 & \textbf{\underline{.26}} & \textbf{.76} & .76 & .51 & \textbf{.83} \\
      & 5& 100 & 98.61  & 88.06 & 62.04 & .78 & .25 & .71 & .95 & .51 & .93 \\
      & 7& 100 & 98.80  & \textbf{88.34}  & \textbf{68.67} & .82 & \textbf{\underline{.26}} & \textbf{\underline{.75}} & \textbf{\underline{.95}} & .50 & \textbf{\underline{.97}} \\
\hline
\multirow{3}{*}{\rotatebox{90}{Phi4}}& 3& 98.26 & 98.76 & 84.77  & 59.88 & .82 & .24 & .69 & .88 & .51 & .80 \\
      & 5& 95.99 & \textbf{99.94}  & 85.70  & \textbf{65.12} & .79 & \textbf{\underline{.26}} & \textbf{\underline{.73}} & .93 & .50 & \textbf{\underline{.94}} \\
      & 7& 98.80 & 99.51  & 85.21  & 65.96 & .82 & \textbf{\underline{.26}} & \textbf{\underline{.75}} & .93 & .51 & \textbf{\underline{.97}} \\
\hline
\multirow{3}{*}{\rotatebox{90}{Qwen2.5}}& 3& 100 & \textbf{99.30}  & 81.55  & 62.79 & .82 & .24 & .70 & .87 & \textbf{\underline{.52}} & .73 \\
         & 5& 99.69 & 98.70  & 86.64  & 62.04 & \textbf{.81} & .24 & .69 & .94 & .51 & .91 \vspace{1pt}\\
         & 7& 95.48 & 99.02 & 85.45 & 60.54 & .82 & .25 & .72 & .93 & \textbf{\underline{.52}} & .96 \vspace{1pt}\\
\hline
\multirow{3}{*}{\rotatebox{90}{Llama3.1}}& 3& 95.64 & 95.64  & 79.46  & 66.28 & .82 & \textbf{\underline{.26}} & .75 & .88 & .51 & .80 \vspace{1pt}\\
         & 5& 85.19 & 85.19 & 71.75 & 56.79 & .79 & \textbf{\underline{.26}} & \textbf{\underline{.73}} & .94 & .50 & \textbf{\underline{.94}} \vspace{1pt}\\
         & 7& 100 & \textbf{99.52} & \textbf{95.18}  & 67.17  & .82 & \textbf{\underline{.26}} & \textbf{\underline{.75}} & .92 & .50 & \textbf{\underline{.97 }}\vspace{1pt}\\
\hline
\multirow{3}{*}{\rotatebox{90}{DS-R1}}& 3& 79.07 & 78.92 & 66.91 & 58.14 & .83 & .25 & .75 & \textbf{\underline{.90}} & .51 & .81 \\
         & 5& 56.17 & 56.17  & 47.40 & 37.04 & .80 & \textbf{\underline{.26}} & \textbf{\underline{.73}} & .93 & .50 & \textbf{\underline{.94 }}\\
         & 7& 41.87 & 41.66  & 34.74  & 30.72 & \textbf{.83} & .25 & .73 & .93 & .51 & \textbf{\underline{.97}} \\
\hline
\multirow{3}{*}{\rotatebox{90}{M-Nemo}}& 3& 98.84 & 98.28  & 84.87  & 67.73 & .81 & .25 & .72 & .88 & .51 & .81 \\
         & 5& 91.36 & 91.30 & 75.65  & 57.10 & .80 & .25 & .72 & .93 & .51 & \textbf{\underline{.94 }}\\
         & 7& 93.67 & 93.37 & 78.16 & 59.94 & .82 & .25 & .73 & .92 & .51 & \textbf{\underline{.97}} \\
\bottomrule
\end{tabular}%
\caption{Performance of \system{} on constraint satisfaction, temporal and structural metrics. We compare the LLM-based scheduler with the deterministic algorithmic scheduler against the monolithic, agentic baselines. $T_m$= Temporal meal score, $T_a$= Temporal attraction score,  $\tilde{T}_a$=normalized temporal attraction score (introduced in ~\system{}), $S_s$=spatial score, $S_p$=persona alignment score, and $S_o$=ordering score. DS-R1=DepSeek-R1, M-Nemo=Mistral-Nemo. }
\label{tab:tripcraft_results}
\end{table}

\section{Experimental Results and Analysis}

This section evaluates \system{}'s performance. We first report benchmark results on the original TripCraft metrics, comparing our agentic framework against monolithic and agentic baselines. We then analyze the impact of integrating review-derived ``Pros'' and ``Cons'' using our review-grounded evaluation metrics.

\subsection{Benchmark Performance on constraint satisfaction, and temporal, structural metrics}

Table~\ref{tab:tripcraft_complete_results} reports the performance of the multi-agent framework without review integration. 
We compare our two scheduling backends (LLM-based scheduler and a deterministic algorithmic scheduler) against the baselines monolithic and agentic prompting approach established in the original TripCraft and ATLAS benchmark. 

These results demonstrate the effectiveness of our agentic decomposition: by breaking the problem down, even smaller open-source models (such as Qwen 2.5 and Phi-4) can maintain high feasibility across multiple trip durations, compared to monolithic approaches that suffer from reasoning overload and outperform the agentic baseline by a significant margin.  

\subsection{Impact of Review Integration}

To evaluate the effect of incorporating review-derived ``Pros'' and ``Cons'', we compare itinerary quality with and without review integration using our proposed review-grounded evaluation metrics. Table~\ref{tab:review_results} reports the results across models and trip durations. The results show that incorporating these qualitative attributes significantly improves persona alignment. 
These findings suggest that integrating user-generated reviews allows the planner to select entities that better reflect real traveler experiences and specific user preferences.

\begin{table}[h]
\centering
\scriptsize
\setlength{\tabcolsep}{1.5pt}
% \resizebox{\textwidth}{!}{%
\begin{tabular}{lc cccc cccc}
\toprule
\multirow{2}{*}{\textbf{}} & \multirow{2}{*}{\textbf{Dur}} &
% \textbf{Constraint} &
\multicolumn{4}{c}{\textbf{LLM Scheduler}} & \multicolumn{4}{c}{\textbf{Deterministic Algo Scheduler}} \\
% \cmidrule(lr){3-3} 
\cmidrule(l){3-6} \cmidrule(l){7-10}
& &
% \textbf{FPR (\%)} &
\textbf{Win (\%)} &
\textbf{Persona} &
\textbf{Exp} &
\textbf{Sat} & \textbf{Win (\%)} &
\textbf{Persona} &
\textbf{Exp} &
\textbf{Sat} \\
% \midrule
% \multicolumn{6}{l}{\textit{\textbf{\system{} (Ours): LLM Planner Pipeline}}} \\
\midrule
\multirow{3}{*}{\rotatebox{90}{GPT5}}
& 3 & 80.52 & 5.09 & 6.95 & 6.26 & 83.08 & 5.14 & 6.95 & 6.31 \\
& 5 & 69.75 & 4.87 & 6.61 & 5.87 & 80.00 & 5.10 & 6.64 & 6.04 \\
& 7 & 79.82 & 4.70 & 6.57 & 5.82 & 77.52 & 4.68 & 6.46 & 5.75 \\
\hline
\multirow{3}{*}{\rotatebox{90}{Phi4}}
& 3 & 73.29 & 4.93 & 6.75 & 6.03 & 73.81 & 4.83 & 6.78 & 6.01\\
& 5 & 71.68 & 5.32 & 6.40 & 5.92& 74.83 & 5.04 & 6.48 & 5.89 \\
& 7 & 80.13 & 5.01 & 6.54 & 5.93 & 83.79 & 5.02 & 6.59 & 5.95 \\
\hline
\multirow{3}{*}{\rotatebox{90}{Qwen2.5}}
& 3 & 80.42 & 4.82 & 6.85 & 6.10 & 79.01 & 4.75 & 7.02 & 6.16 \\
& 5 & 71.53 & 4.99 & 6.44 & 5.84 & 68.01 & 4.92 & 6.64 & 5.92\vspace{1pt}\\
& 7 & 77.98 & 4.88 & 6.57 & 5.89 & 83.54 & 5.03 & 6.74 & 6.11\vspace{1pt} \\
\hline
\multirow{3}{*}{\rotatebox{90}{Llama3.1}}
& 3 & 75.67 & 4.80 & 6.80 & 5.96 & 71.66 & 4.65 & 6.70 & 5.81 \\
& 5 & 75.96 & 5.27 & 6.45 & 5.93 & 74.86 & 5.17 & 6.52 & 5.92\vspace{1pt}\\
& 7 & 80.19 & 5.07 & 6.56 & 5.98 & 84.13 & 4.95 & 6.57 & 5.93\vspace{1pt} \\
\hline
\multirow{3}{*}{\rotatebox{90}{DS-R1}}
& 3 & 76.26 & 4.99 & 6.82 & 6.09 & 67.73 & 4.95 & 6.79 & 6.00\\
& 5 & 79.70 & 5.52 & 6.71 & 6.28 & 77.78 & 5.47 & 6.78 & 6.31 \\
& 7 & 78.38 & 5.27 & 6.69 & 6.16 & 87.36 & 5.43 & 6.80 & 6.24 \\
\hline
\multirow{3}{*}{\rotatebox{90}{M-Nemo}}
& 3 & 66.97 & 4.64 & 6.51 & 5.68& 62.80 & 4.49 & 6.34 & 5.45 \\
& 5 & 63.25 & 4.86 & 6.32 & 5.69 & 55.41 & 4.73 & 6.22 & 5.46\\
& 7 & 66.32 & 4.59 & 6.27 & 5.55 & 61.09 & 4.41 & 6.16 & 5.36\\

\bottomrule
\end{tabular}%
\caption{RGPA metrics: Impact of review integration on itinerary quality across both the LLM Scheduler Pipeline and the Deterministic Algorithmic Scheduler in \system{}. Win Rate denotes \% of pairwise comparisons in which the review-grounded itinerary was preferred over the corresponding non-review itinerary by the LLM evaluator. Persona, Experience, and Satisfaction correspond to averaged judge-assigned scores on a 1--10 scale.
}
\label{tab:review_results}
\end{table}

\begin{table}[t]
\centering
\caption{Evaluation of RGPA using Qwen3-32B as an independent LLM judge.}
\label{tab:cross_judge}
\small
\setlength{\tabcolsep}{3.5pt}
\begin{tabular}{lcccc}
\toprule
\textbf{Plan} & \textbf{Win} & \textbf{Persona} & \textbf{Exp.} & \textbf{Sat.} \\
\midrule
3-day & 81.00 & 8.39 & 7.89 & 7.82 \\
5-day & 65.49 & 8.48 & 7.76 & 7.75 \\
7-day & 66.60 & 8.23 & 7.47 & 7.44 \\
\bottomrule
\end{tabular}
\end{table}

\subsection{Robustness and Quality Validation}

\paragraph{Inter-judge robustness.}
We also evaluated RGPA using Qwen3-32B as an independent LLM judge. As shown in Table~\ref{tab:cross_judge}, the review-grounded itineraries achieve Win Rates of 81.00\%, 65.49\%, and 66.60\% for 3-, 5-, and 7-day plans, respectively. These results are consistent with the GPT-5 evaluation in Table~\ref{tab:review_results}, indicating that the observed improvements are not specific to a particular judge. Although the absolute Persona, Experience, and Satisfaction scores vary across judges, likely due to differences in scoring calibration, the relative preference for review-grounded itineraries remains consistent. Therefore, \textbf{Win Rate serves as a robust cross-judge indicator}, while the GPT-5-based absolute scores are used for the detailed analysis of RGPA.

\paragraph{Human validation.}
We manually evaluated 40 itinerary pairs from the 3-day setting. The review-grounded itinerary was judged equal to or better than the baseline in 31 cases (77.5\%), closely matching the 80.52\% GPT-5 Win Rate. The improvements primarily reflected better persona-aware entity selection, such as culturally relevant attractions, highly rated restaurants, and premium venues for travelers with luxury-oriented preferences. Thus, the observed gains extend beyond superficial differences in textual presentation and reflect substantive improvements in the composition of the generated itineraries. Collectively, the cross-judge and human evaluations provide evidence that RGPA's improvements are robust to the choice of evaluator and reflect meaningful improvements in travel-planning quality.

\paragraph{Dataset-level analysis.}
We analyzed the extracted Pros and Cons across 2,369 accommodations, 3,791 restaurants, and 4,897 attractions. Empty-Pro rates remained below 1\% across all domains, while empty-Cons rates were higher, particularly for accommodations (53.14\%). This reflects the predominantly positive nature of many accommodation reviews. Table~\ref{tab:summary_quality} further shows high extraction precision and sentiment accuracy, particularly for positive signals.

\begin{table}[t]
\centering
\caption{Dataset-level statistics of extracted Pros and Cons.}
\label{tab:dataset_analysis}
\small
\setlength{\tabcolsep}{3pt}
\begin{tabular}{lccc}
\toprule
\textbf{Statistic} & \textbf{Acc.} & \textbf{Rest.} & \textbf{Attr.}\\
\midrule
Entities & 2,369 & 3,791 & 4,897\\
Reviews/entity & 7.65 & 11.22 & 10.41\\
Pros/entity & 5.09 & 4.98 & 9.47\\
Cons/entity & 2.00 & 4.16 & 7.11\\
Empty Pros (\%) & 0.08 & 0.26 & 0.76\\
Empty Cons (\%) & 53.14 & 16.42 & 29.39\\
\bottomrule
\end{tabular}
\end{table}

\begin{table}[t]
\centering
\caption{Quality of extracted Pros and Cons using Qwen3-32B.}
\label{tab:summary_quality}
\small
\setlength{\tabcolsep}{3pt}
\begin{tabular}{lccc}
\toprule
\textbf{Metric} & \textbf{Acc.} & \textbf{Rest.} & \textbf{Attr.}\\
\midrule
Precision (Pro) & 98.00 & 96.40 & 99.57\\
Precision (Con) & 75.60 & 73.10 & 96.07\\
Sentiment (Pro) & 100.00 & 100.00 & 99.57\\
Sentiment (Con) & 91.30 & 97.80 & 96.07\\
\bottomrule
\end{tabular}
\end{table}

\begin{tcolorbox}[
    colback=yellow!10,
    colframe=yellow!50!black,
    boxrule=0.4pt,
    arc=1pt,
    left=4pt,
    right=4pt,
    top=4pt,
    bottom=4pt
]
\small

\textbf{Example 1: Accommodation.}
\textit{Island Home, 1 BR Suite w/Harbor View (Vineyard Haven).}

\textbf{Review:} ``A lot of old and mismatched decor. The shower curtain
in the tub is old ... the whole place is on a slant ... the doorways are
kind of low and I smashed my head a few times.''

\textbf{Qwen-extracted Cons:}
\begin{itemize}
    \item Old and mismatched decor.
    \item Old shower curtain that is not frequently replaced.
    \item Building is on a slant.
    \item Low doorways may cause head injuries for taller guests.
\end{itemize}

\vspace{2pt}

\textbf{Example 2: Restaurant.}
\textit{2M Smokehouse \& Catering (San Antonio, TX).}

\textbf{Review:} ``The standout was the brisket. So delish, and you could
cut it with a fork.'' / ``Amazing brisket ... Ribs were also outstanding.
The pork and turkey were darn good as well ... The street corn [was]
yummy as were the beans and slaw.'' / ``Locals told us this was the place
to go.''

\textbf{Qwen-extracted Pros:}
\begin{itemize}
    \item Amazing BBQ with standout, tender brisket.
    \item Excellent ribs, pork, and turkey.
    \item Delicious sides including street corn, beans, and slaw.
    \item Highly recommended for an authentic local BBQ experience.
\end{itemize}

\end{tcolorbox}

Overall, the results indicate that RGPA's review-derived experiential signals are reliable, well grounded, and robust across both automated and human evaluation.

\subsection{Qualitative Analysis of Scheduling Failures}

As shown in Table~\ref{tab:tripcraft_results}, generative LLM-based scheduling struggles with strict constraint satisfaction, particularly as trip duration increases (e.g., the 0\% Pass Rate for Qwen 2.5 on 7-day queries). 

We explicitly designed this comparison to demonstrate that while LLMs excel at semantic ranking and entity selection, they suffer from severe reasoning drift when tasked with dense, symbolic spatio-temporal math.

A qualitative analysis of the generated itineraries highlights several recurring hallucination patterns in the generative pathway that the deterministic scheduler successfully mitigates:

    (1) Duration Hallucinations: Generative models frequently output zero-duration stays. For instance, in our generated samples, Qwen 2.5 explicitly scheduled an accommodation stay from ``10:42 to 10:42'' and later from ``12:00 to 12:00''.
    (2) Chronological Inversions: Generative schedulers struggle to maintain a coherent 24-hour clock logic. In a generated 3-day itinerary, an LLM scheduled a lunch reservation at 3:30 AM (``visit from 03:30 to 04:30''), critically violating the commonsense constraint for daytime dining windows.
    (3) Transit Buffer Violations: LLMs often ignore mandatory transit times between locations. In the same generated schedule, the LLM booked a restaurant visit ending at 13:00 while completely neglecting the 30-minute travel buffer required before the next activity.

By contrast, the deterministic algorithmic scheduler rigorously maps the ``Pros'' and ``Cons'' selected by the semantic agents onto a 
mathematically verified skeleton, consistently enforcing meal gaps, transit buffers, and logical ordering. A detailed failure analysis is provided in App.~\ref{app:scheduling_failures}.

\section{Discussion}

\noindent\textbf{Impact of Agentic Decomposition.}
Decomposing the itinerary generation task into specialized agents significantly improves constraint satisfaction compared to monolithic prompting approaches. By isolating reasoning to localized contexts (e.g., evaluating only restaurants within a single destination city), our framework significantly reduces reasoning overload and prevents the hallucination of entities. This bounded-context design is the primary driver that enables efficient, open-source models (such as Qwen~2.5 and Phi-4) to achieve planning stability and constraint adherence competitive with massive proprietary models.

\noindent\textbf{Effect of Review Integration.}
Relying solely on structural databases limits a system's ability to understand the qualitative realities of travel. Incorporating structured ``Pros'' and ``Cons'' directly improves experiential metrics such as RGPA. More importantly, these qualitative attributes resolve the semantic rigidity of traditional databases. By grounding decisions in actual user feedback, the agents can successfully map nuanced, subjective persona requests (e.g., a desire for a ``cozy'' atmosphere) to entities that consistently deliver those experiences, while simultaneously penalizing locations flagged for hygiene or safety risks.

\noindent\textbf{LLM vs. Algorithmic Scheduling.}
The comparison between the LLM-based scheduler pipeline and the deterministic Algorithmic Scheduler highlights a crucial architectural trade-off. While the LLM-based scheduler offers high semantic flexibility, the deterministic scheduler guarantees strict temporal and budget constraint satisfaction while significantly reducing computational overhead (i.e., fewer LLM calls). This validates our core hypothesis: a hybrid neuro-symbolic architecture, where LLMs handle the 
entity selection and ranking, while a deterministic algorithm handles the rigid mathematics of spatio-temporal scheduling, provides the most robust and reliable path forward for complex automated planning.

\section{Conclusion}

In this paper, we introduced \system{}, a neuro-symbolic framework for spatio-temporal travel planning. The proposed system decomposes itinerary generation into specialized agents responsible for accommodations, transportation, dining, attractions, and events, coordinated through a central Global Orchestrator. We further integrate review-derived ``Pros'' and ``Cons'' into the planning process and propose the RGPA metric to comprehensively evaluate experience quality, personalization, and risk avoidance. Experimental results on the augmented TripCraft benchmark demonstrate that the proposed framework improves constraint satisfaction and planning reliability while enabling effective use of both proprietary and open-source language models. These findings highlight the potential of agentic architectures for complex planning tasks.

\section{Limitations}

Although the proposed framework significantly improves constraint satisfaction and experiential quality, several limitations remain. First, the distributed and sequential nature of the multi-agent architecture introduces additional inference latency and orchestration overhead compared to single-pass monolithic generation, presenting a trade-off between semantic reasoning quality and real-time execution speed. Second, the extraction of ``Pros'' and ``Cons'' relies on the automated processing of user-generated text, making the pipeline potentially sensitive to noise, review manipulation (e.g., review bombing), or domain biases inherent in the scraped datasets. Finally, our evaluation is conducted on the TripCraft benchmark, which primarily encompasses U.S. travel scenarios and structured transit networks; consequently, it may not fully capture the complexities of global travel planning in regions with less formalized tourism infrastructure.

\section{Ethical Considerations}
The user reviews collected to augment the TripCraft benchmark were obtained from publicly available platforms. We did not directly recruit or interact with the individuals who authored these reviews, and therefore individual consent was not obtained. To ensure user privacy, all personally identifiable information (PII) and reviewer usernames were stripped during the preprocessing phase. The NLP pipeline aggregates these qualitative attributes strictly at the entity level, preventing the profiling of individual reviewers. Furthermore, we acknowledge that LLMs and user reviews can encode societal biases; our explicit extraction of safety-related ``Cons'' attempts to mitigate the recommendation of unsafe entities, though we recognize that automated safety filtering is not infallible. AI-assisted tools were used during the preparation of this work for language refinement and polishing. All research decisions, methodology, data curation, experiments, analyses, and conclusions were performed and verified by the authors. 

\section*{Acknowledgments}
This research was partially supported by the Technology Innovation Hub (TIH), IIT Tirupati (IITTNiF/TPD/2024-25/P16). We sincerely thank Soutrik Das from IIT Bhubaneswar for his assistance with baseline implementation and quality checking. Finally, we thank the anonymous reviewers for their valuable comments and constructive feedback, which helped improve this work.

\bibliography{custom}

@inproceedings{xie2024travelplanner,
  author       = {Jian Xie and
                  Kai Zhang and
                  Jiangjie Chen and
                  Tinghui Zhu and
                  Renze Lou and
                  Yuandong Tian and
                  Yanghua Xiao and
                  Yu Su},
  editor       = {Ruslan Salakhutdinov and
                  Zico Kolter and
                  Katherine A. Heller and
                  Adrian Weller and
                  Nuria Oliver and
                  Jonathan Scarlett and
                  Felix Berkenkamp},
  title        = {TravelPlanner: {A} Benchmark for Real-World Planning with Language
                  Agents},
  booktitle    = {Forty-first International Conference on Machine Learning, {ICML} 2024,
                  Vienna, Austria, July 21-27, 2024},
  series       = {Proceedings of Machine Learning Research},
  volume       = {235},
  pages        = {54590--54613},
  publisher    = {{PMLR} / OpenReview.net},
  year         = {2024},
  url          = {https://proceedings.mlr.press/v235/xie24j.html},
  bibsource    = {dblp computer science bibliography, https://dblp.org}
}

@inproceedings{karmakar-etal-2026-triptide,
    title = "{T}rip{T}ide: A Benchmark for Adaptive Travel Planning under Disruptions",
    author = "Karmakar, Priyanshu  and
      Chaudhuri, Soumyabrata  and
      Mallick, Shubhojit  and
      Gupta, Manish  and
      Jana, Abhik  and
      Ghosh, Shreya",
    editor = "Liakata, Maria  and
      Moreira, Viviane P.  and
      Zhang, Jiajun  and
      Jurgens, David",
    booktitle = "Findings of the {A}ssociation for {C}omputational {L}inguistics: {ACL} 2026",
    month = jul,
    year = "2026",
    address = "San Diego, California, United States",
    publisher = "Association for Computational Linguistics",
    url = "https://aclanthology.org/2026.findings-acl.2002/",
    doi = "10.18653/v1/2026.findings-acl.2002",
    pages = "40269--40292",
    ISBN = "979-8-89176-395-1"
}

@inproceedings{chaudhuri2025tripcraft,
  title={Tripcraft: A benchmark for spatio-temporally fine grained travel planning},
  author={Chaudhuri, Soumyabrata and Purkar, Pranav and Raghav, Ritwik and Mallick, Shubhojit and Gupta, Manish and Jana, Abhik and Ghosh, Shreya},
  booktitle={Proceedings of the 63rd Annual Meeting of the Association for Computational Linguistics (Volume 1: Long Papers)},
  pages={17035--17064},
  year={2025}
}

@inproceedings{hao2025formalplanning,
  title={Large language models can solve real-world planning rigorously with formal verification tools},
  author={Hao, Yilun and Chen, Yongchao and Zhang, Yang and Fan, Chuchu},
  booktitle={Proceedings of the 2025 Conference of the Nations of the Americas Chapter of the Association for Computational Linguistics: Human Language Technologies (Volume 1: Long Papers)},
  pages={3434--3483},
  year={2025}
}

@inproceedings{wei2022chain,
  author       = {Jason Wei and
                  Xuezhi Wang and
                  Dale Schuurmans and
                  Maarten Bosma and
                  Brian Ichter and
                  Fei Xia and
                  Ed H. Chi and
                  Quoc V. Le and
                  Denny Zhou},
  editor       = {Sanmi Koyejo and
                  S. Mohamed and
                  A. Agarwal and
                  Danielle Belgrave and
                  K. Cho and
                  A. Oh},
  title        = {Chain-of-Thought Prompting Elicits Reasoning in Large Language Models},
  booktitle    = {Advances in Neural Information Processing Systems 35: Annual Conference
                  on Neural Information Processing Systems 2022, NeurIPS 2022, New Orleans,
                  LA, USA, November 28 - December 9, 2022},
  year         = {2022},
  url          = {http://papers.nips.cc/paper\_files/paper/2022/hash/9d5609613524ecf4f15af0f7b31abca4-Abstract-Conference.html},
  bibsource    = {dblp computer science bibliography, https://dblp.org}
}

@inproceedings{yao2023react,
  author       = {Shunyu Yao and
                  Jeffrey Zhao and
                  Dian Yu and
                  Nan Du and
                  Izhak Shafran and
                  Karthik R. Narasimhan and
                  Yuan Cao},
  title        = {ReAct: Synergizing Reasoning and Acting in Language Models},
  booktitle    = {The Eleventh International Conference on Learning Representations,
                  {ICLR} 2023, Kigali, Rwanda, May 1-5, 2023},
  publisher    = {OpenReview.net},
  year         = {2023},
  url          = {https://openreview.net/forum?id=WE\_vluYUL-X},
  bibsource    = {dblp computer science bibliography, https://dblp.org}
}

@incollection{BT18,
   author = {Clark Barrett and Cesare Tinelli},
   editor = {Edmund M. Clarke and Thomas A. Henzinger and Helmut Veith and
	Roderick Bloem},
   title = {Satisfiability Modulo Theories},
   booktitle = {Handbook of Model Checking},
   pages = {305--343},
   publisher = {Springer International Publishing},
   year = {2018},
   isbn = {978-3-319-10575-8},
   doi = {10.1007/978-3-319-10575-8_11},
   url = {http://theory.stanford.edu/~barrett/pubs/BT18.pdf}
}

@inproceedings{yao2023tree,
  author       = {Shunyu Yao and
                  Dian Yu and
                  Jeffrey Zhao and
                  Izhak Shafran and
                  Tom Griffiths and
                  Yuan Cao and
                  Karthik Narasimhan},
  editor       = {Alice Oh and
                  Tristan Naumann and
                  Amir Globerson and
                  Kate Saenko and
                  Moritz Hardt and
                  Sergey Levine},
  title        = {Tree of Thoughts: Deliberate Problem Solving with Large Language Models},
  booktitle    = {Advances in Neural Information Processing Systems 36: Annual Conference
                  on Neural Information Processing Systems 2023, NeurIPS 2023, New Orleans,
                  LA, USA, December 10 - 16, 2023},
  year         = {2023},
  url          = {http://papers.nips.cc/paper\_files/paper/2023/hash/271db9922b8d1f4dd7aaef84ed5ac703-Abstract-Conference.html},
  bibsource    = {dblp computer science bibliography, https://dblp.org}
}

@article{shinn2024reflexion,
  title={Reflexion: Language agents with verbal reinforcement learning},
  author={Shinn, Noah and Cassano, Federico and Gopinath, Ashwin and Narasimhan, Karthik and Yao, Shunyu},
  journal={Advances in neural information processing systems},
  volume={36},
  pages={8634--8652},
  year={2023}
}

@inproceedings{hong2024metagpt,
 author = {Hong, Sirui and Zhuge, Mingchen and Chen, Jonathan and Zheng, Xiawu and Cheng, Yuheng and Wang, Jinlin and Zhang, Ceyao and wang, zili and Yau, Steven and Lin, Zijuan and Zhou, Liyang and Ran, Chenyu and Xiao, Lingfeng and Wu, Chenglin and Schmidhuber, J\"{u}rgen},
 booktitle = {International Conference on Learning Representations},
 editor = {B. Kim and Y. Yue and S. Chaudhuri and K. Fragkiadaki and M. Khan and Y. Sun},
 pages = {23247--23275},
 title = {MetaGPT: Meta Programming for A Multi-Agent Collaborative Framework},
 url = {https://proceedings.iclr.cc/paper_files/paper/2024/file/6507b115562bb0a305f1958ccc87355a-Paper-Conference.pdf},
 volume = {2024},
 year = {2024}
}

@article{li2023camel,
  title={Camel: Communicative agents for "mind" exploration of large language model society},
  author={Li, Guohao and Hammoud, Hasan and Itani, Hani and Khizbullin, Dmitrii and Ghanem, Bernard},
  journal={Advances in neural information processing systems},
  volume={36},
  pages={51991--52008},
  year={2023}
}

@inproceedings{valmeekam2023large,
  title={Large language models still can't plan (a benchmark for LLMs on planning and reasoning about change)},
  author={Valmeekam, Karthik and Olmo, Alberto and Sreedharan, Sarath and Kambhampati, Subbarao},
  booktitle={NeurIPS 2022 Foundation Models for Decision Making Workshop},
  year={2022}
}

@inproceedings{ni2025tprag,
    title = "{TP}-{RAG}: Benchmarking Retrieval-Augmented Large Language Model Agents for Spatiotemporal-Aware Travel Planning",
    author = "Ni, Hang  and
      Liu, Fan  and
      Ma, Xinyu  and
      Su, Lixin  and
      Wang, Shuaiqiang  and
      Yin, Dawei  and
      Xiong, Hui  and
      Liu, Hao",
    editor = "Christodoulopoulos, Christos  and
      Chakraborty, Tanmoy  and
      Rose, Carolyn  and
      Peng, Violet",
    booktitle = "Proceedings of the 2025 Conference on Empirical Methods in Natural Language Processing",
    month = nov,
    year = "2025",
    address = "Suzhou, China",
    publisher = "Association for Computational Linguistics",
    url = "https://aclanthology.org/2025.emnlp-main.626/",
    doi = "10.18653/v1/2025.emnlp-main.626",
    pages = "12392--12418",
    ISBN = "979-8-89176-332-6"
}

@inproceedings{ahn2022can,
  title={Do as i can, not as i say: Grounding language in robotic affordances},
  author={Ahn, Michael and Brohan, Anthony and Brown, Noah and Chebotar, Yevgen and Cortes, Omar and David, Byron and Finn, Chelsea and Fu, Chuyuan and Gopalakrishnan, Keerthana and Hausman, Karol and others},
  booktitle={Conference on robot learning},
  pages={287--318},
  year={2023},
  organization={Pmlr}
}

@article{wang2023voyager,
  title={Voyager: An open-ended embodied agent with large language models},
  author={Wang, Guanzhi and Xie, Yuqi and Jiang, Yunfan and Mandlekar, Ajay and Xiao, Chaowei and Zhu, Yuke and Fan, Linxi and Anandkumar, Anima},
  journal={arXiv preprint arXiv:2305.16291},
  year={2023}
}

@article{liu2023llmpddl,
  title={Llm+ p: Empowering large language models with optimal planning proficiency},
  author={Liu, Bo and Jiang, Yuqian and Zhang, Xiaohan and Liu, Qiang and Zhang, Shiqi and Biswas, Joydeep and Stone, Peter},
  journal={arXiv preprint arXiv:2304.11477},
  year={2023}
}

@article{guan2023leveraging,
  title={Leveraging pre-trained large language models to construct and utilize world models for model-based task planning},
  author={Guan, Lin and Valmeekam, Karthik and Sreedharan, Sarath and Kambhampati, Subbarao},
  journal={Advances in Neural Information Processing Systems},
  volume={36},
  pages={79081--79094},
  year={2023}
}

@inproceedings{pontiki2014semeval,
    title = "{S}em{E}val-2014 Task 4: Aspect Based Sentiment Analysis",
    author = "Pontiki, Maria  and
      Galanis, Dimitris  and
      Pavlopoulos, John  and
      Papageorgiou, Harris  and
      Androutsopoulos, Ion  and
      Manandhar, Suresh",
    editor = "Nakov, Preslav  and
      Zesch, Torsten",
    booktitle = "Proceedings of the 8th International Workshop on Semantic Evaluation ({S}em{E}val 2014)",
    month = aug,
    year = "2014",
    address = "Dublin, Ireland",
    publisher = "Association for Computational Linguistics",
    url = "https://aclanthology.org/S14-2004/",
    doi = "10.3115/v1/S14-2004",
    pages = "27--35"
}

@inproceedings{zheng2017joint,
  author       = {Lei Zheng and
                  Vahid Noroozi and
                  Philip S. Yu},
  editor       = {Maarten de Rijke and
                  Milad Shokouhi and
                  Andrew Tomkins and
                  Min Zhang},
  title        = {Joint Deep Modeling of Users and Items Using Reviews for Recommendation},
  booktitle    = {Proceedings of the Tenth {ACM} International Conference on Web Search
                  and Data Mining, {WSDM} 2017, Cambridge, United Kingdom, February
                  6-10, 2017},
  pages        = {425--434},
  publisher    = {{ACM}},
  year         = {2017},
  url          = {https://doi.org/10.1145/3018661.3018665},
  doi          = {10.1145/3018661.3018665},
  bibsource    = {dblp computer science bibliography, https://dblp.org}
}

@inproceedings{10.1145/3178876.3186070,
author = {Chen, Chong and Zhang, Min and Liu, Yiqun and Ma, Shaoping},
title = {Neural Attentional Rating Regression with Review-level Explanations},
year = {2018},
isbn = {9781450356398},
publisher = {International World Wide Web Conferences Steering Committee},
address = {Republic and Canton of Geneva, CHE},
url = {https://doi.org/10.1145/3178876.3186070},
doi = {10.1145/3178876.3186070},
booktitle = {Proceedings of the 2018 World Wide Web Conference},
pages = {1583–1592},
numpages = {10},
location = {Lyon, France},
series = {WWW '18}
}

@article{wei2025learning,
  title={Learning to Shop Like Humans: A Review-driven Retrieval-Augmented Recommendation Framework with LLMs},
  author={Wei, Kaiwen and Gao, Jinpeng and Zhong, Jiang and Yang, Yuming and Lv, Fengmao and Li, Zhenyang},
  journal={arXiv preprint arXiv:2509.00698},
  year={2025}
}

@inproceedings{zheng2023judging,
  author       = {Lianmin Zheng and
                  Wei{-}Lin Chiang and
                  Ying Sheng and
                  Siyuan Zhuang and
                  Zhanghao Wu and
                  Yonghao Zhuang and
                  Zi Lin and
                  Zhuohan Li and
                  Dacheng Li and
                  Eric P. Xing and
                  Hao Zhang and
                  Joseph E. Gonzalez and
                  Ion Stoica},
  editor       = {Alice Oh and
                  Tristan Naumann and
                  Amir Globerson and
                  Kate Saenko and
                  Moritz Hardt and
                  Sergey Levine},
  title        = {Judging LLM-as-a-Judge with MT-Bench and Chatbot Arena},
  booktitle    = {Advances in Neural Information Processing Systems 36: Annual Conference
                  on Neural Information Processing Systems 2023, NeurIPS 2023, New Orleans,
                  LA, USA, December 10 - 16, 2023},
  year         = {2023},
  url          = {http://papers.nips.cc/paper\_files/paper/2023/hash/91f18a1287b398d378ef22505bf41832-Abstract-Datasets\_and\_Benchmarks.html},
  bibsource    = {dblp computer science bibliography, https://dblp.org}
}

@misc{mistral2024nemo,
  author       = {Mistral AI},
  title        = {{Mistral NeMo}},
  year         = {2024},
  howpublished = {\url{https://mistral.ai/news/mistral-nemo}},
  note         = {Accessed: 2024}
}

@inproceedings{choi2026atlas,
  title={ATLAS: Constraints-aware multi-agent collaboration for real-world travel planning},
  author={Choi, Jihye and Yoon, Jinsung and Chen, Jiefeng and Jha, Somesh and Pfister, Tomas},
  booktitle={International Conference on Learning Representations},
  year={2026}
}

@article{Guo_2025,
   title={DeepSeek-R1 incentivizes reasoning in LLMs through reinforcement learning},
   volume={645},
   ISSN={1476-4687},
   url={http://dx.doi.org/10.1038/s41586-025-09422-z},
   DOI={10.1038/s41586-025-09422-z},
   number={8081},
   journal={Nature},
   publisher={Springer Science and Business Media LLC},
   author={Guo, Daya and Yang, Dejian and Zhang, Haowei and Song, Junxiao and Wang, Peiyi and Zhu, Qihao and Xu, Runxin and Zhang, Ruoyu and Ma, Shirong and Bi, Xiao and Zhang, Xiaokang and Yu, Xingkai and Wu, Yu and Wu, Z. F. and Gou, Zhibin and Shao, Zhihong and Li, Zhuoshu and Gao, Ziyi and Liu, Aixin and Xue, Bing and Wang, Bingxuan and Wu, Bochao and Feng, Bei and Lu, Chengda and Zhao, Chenggang and Deng, Chengqi and Ruan, Chong and Dai, Damai and Chen, Deli and Ji, Dongjie and Li, Erhang and Lin, Fangyun and Dai, Fucong and Luo, Fuli and Hao, Guangbo and Chen, Guanting and Li, Guowei and Zhang, H. and Xu, Hanwei and Ding, Honghui and Gao, Huazuo and Qu, Hui and Li, Hui and Guo, Jianzhong and Li, Jiashi and Chen, Jingchang and Yuan, Jingyang and Tu, Jinhao and Qiu, Junjie and Li, Junlong and Cai, J. L. and Ni, Jiaqi and Liang, Jian and Chen, Jin and Dong, Kai and Hu, Kai and You, Kaichao and Gao, Kaige and Guan, Kang and Huang, Kexin and Yu, Kuai and Wang, Lean and Zhang, Lecong and Zhao, Liang and Wang, Litong and Zhang, Liyue and Xu, Lei and Xia, Leyi and Zhang, Mingchuan and Zhang, Minghua and Tang, Minghui and Zhou, Mingxu and Li, Meng and Wang, Miaojun and Li, Mingming and Tian, Ning and Huang, Panpan and Zhang, Peng and Wang, Qiancheng and Chen, Qinyu and Du, Qiushi and Ge, Ruiqi and Zhang, Ruisong and Pan, Ruizhe and Wang, Runji and Chen, R. J. and Jin, R. L. and Chen, Ruyi and Lu, Shanghao and Zhou, Shangyan and Chen, Shanhuang and Ye, Shengfeng and Wang, Shiyu and Yu, Shuiping and Zhou, Shunfeng and Pan, Shuting and Li, S. S. and Zhou, Shuang and Wu, Shaoqing and Yun, Tao and Pei, Tian and Sun, Tianyu and Wang, T. and Zeng, Wangding and Liu, Wen and Liang, Wenfeng and Gao, Wenjun and Yu, Wenqin and Zhang, Wentao and Xiao, W. L. and An, Wei and Liu, Xiaodong and Wang, Xiaohan and Chen, Xiaokang and Nie, Xiaotao and Cheng, Xin and Liu, Xin and Xie, Xin and Liu, Xingchao and Yang, Xinyu and Li, Xinyuan and Su, Xuecheng and Lin, Xuheng and Li, X. Q. and Jin, Xiangyue and Shen, Xiaojin and Chen, Xiaosha and Sun, Xiaowen and Wang, Xiaoxiang and Song, Xinnan and Zhou, Xinyi and Wang, Xianzu and Shan, Xinxia and Li, Y. K. and Wang, Y. Q. and Wei, Y. X. and Zhang, Yang and Xu, Yanhong and Li, Yao and Zhao, Yao and Sun, Yaofeng and Wang, Yaohui and Yu, Yi and Zhang, Yichao and Shi, Yifan and Xiong, Yiliang and He, Ying and Piao, Yishi and Wang, Yisong and Tan, Yixuan and Ma, Yiyang and Liu, Yiyuan and Guo, Yongqiang and Ou, Yuan and Wang, Yuduan and Gong, Yue and Zou, Yuheng and He, Yujia and Xiong, Yunfan and Luo, Yuxiang and You, Yuxiang and Liu, Yuxuan and Zhou, Yuyang and Zhu, Y. X. and Huang, Yanping and Li, Yaohui and Zheng, Yi and Zhu, Yuchen and Ma, Yunxian and Tang, Ying and Zha, Yukun and Yan, Yuting and Ren, Z. Z. and Ren, Zehui and Sha, Zhangli and Fu, Zhe and Xu, Zhean and Xie, Zhenda and Zhang, Zhengyan and Hao, Zhewen and Ma, Zhicheng and Yan, Zhigang and Wu, Zhiyu and Gu, Zihui and Zhu, Zijia and Liu, Zijun and Li, Zilin and Xie, Ziwei and Song, Ziyang and Pan, Zizheng and Huang, Zhen and Xu, Zhipeng and Zhang, Zhongyu and Zhang, Zhen},
   year={2025},
   month=Sept, pages={633–638} }

@misc{qwen2025qwen25technicalreport,
      title={Qwen2.5 Technical Report}, 
      author={Qwen and : and An Yang and Baosong Yang and Beichen Zhang and Binyuan Hui and Bo Zheng and Bowen Yu and Chengyuan Li and Dayiheng Liu and Fei Huang and Haoran Wei and Huan Lin and Jian Yang and Jianhong Tu and Jianwei Zhang and Jianxin Yang and Jiaxi Yang and Jingren Zhou and Junyang Lin and Kai Dang and Keming Lu and Keqin Bao and Kexin Yang and Le Yu and Mei Li and Mingfeng Xue and Pei Zhang and Qin Zhu and Rui Men and Runji Lin and Tianhao Li and Tianyi Tang and Tingyu Xia and Xingzhang Ren and Xuancheng Ren and Yang Fan and Yang Su and Yichang Zhang and Yu Wan and Yuqiong Liu and Zeyu Cui and Zhenru Zhang and Zihan Qiu},
      year={2025},
      eprint={2412.15115},
      archivePrefix={arXiv},
      primaryClass={cs.CL},
      url={https://arxiv.org/abs/2412.15115}, 
}

@misc{microsoft2025phi4minitechnicalreportcompact,
      title={Phi-4-Mini Technical Report: Compact yet Powerful Multimodal Language Models via Mixture-of-LoRAs}, 
      author={Microsoft and : and Abdelrahman Abouelenin and Atabak Ashfaq and Adam Atkinson and Hany Awadalla and Nguyen Bach and Jianmin Bao and Alon Benhaim and Martin Cai and Vishrav Chaudhary and Congcong Chen and Dong Chen and Dongdong Chen and Junkun Chen and Weizhu Chen and Yen-Chun Chen and Yi-ling Chen and Qi Dai and Xiyang Dai and Ruchao Fan and Mei Gao and Min Gao and Amit Garg and Abhishek Goswami and Junheng Hao and Amr Hendy and Yuxuan Hu and Xin Jin and Mahmoud Khademi and Dongwoo Kim and Young Jin Kim and Gina Lee and Jinyu Li and Yunsheng Li and Chen Liang and Xihui Lin and Zeqi Lin and Mengchen Liu and Yang Liu and Gilsinia Lopez and Chong Luo and Piyush Madan and Vadim Mazalov and Arindam Mitra and Ali Mousavi and Anh Nguyen and Jing Pan and Daniel Perez-Becker and Jacob Platin and Thomas Portet and Kai Qiu and Bo Ren and Liliang Ren and Sambuddha Roy and Ning Shang and Yelong Shen and Saksham Singhal and Subhojit Som and Xia Song and Tetyana Sych and Praneetha Vaddamanu and Shuohang Wang and Yiming Wang and Zhenghao Wang and Haibin Wu and Haoran Xu and Weijian Xu and Yifan Yang and Ziyi Yang and Donghan Yu and Ishmam Zabir and Jianwen Zhang and Li Lyna Zhang and Yunan Zhang and Xiren Zhou},
      year={2025},
      eprint={2503.01743},
      archivePrefix={arXiv},
      primaryClass={cs.CL},
      url={https://arxiv.org/abs/2503.01743}, 
}

@misc{grattafiori2024llama3herdmodels,
      title={The Llama 3 Herd of Models}, 
      author={Aaron Grattafiori and Abhimanyu Dubey and Abhinav Jauhri and Abhinav Pandey and Abhishek Kadian and Ahmad Al-Dahle and Aiesha Letman and Akhil Mathur and Alan Schelten and Alex Vaughan and Amy Yang and Angela Fan and Anirudh Goyal and Anthony Hartshorn and Aobo Yang and Archi Mitra and Archie Sravankumar and Artem Korenev and Arthur Hinsvark and Arun Rao and Aston Zhang and Aurelien Rodriguez and Austen Gregerson and Ava Spataru and Baptiste Roziere and Bethany Biron and Binh Tang and Bobbie Chern and Charlotte Caucheteux and Chaya Nayak and Chloe Bi and Chris Marra and Chris McConnell and Christian Keller and Christophe Touret and Chunyang Wu and Corinne Wong and Cristian Canton Ferrer and Cyrus Nikolaidis and Damien Allonsius and Daniel Song and Danielle Pintz and Danny Livshits and Danny Wyatt and David Esiobu and Dhruv Choudhary and Dhruv Mahajan and Diego Garcia-Olano and Diego Perino and Dieuwke Hupkes and Egor Lakomkin and Ehab AlBadawy and Elina Lobanova and Emily Dinan and Eric Michael Smith and Filip Radenovic and Francisco Guzmán and Frank Zhang and Gabriel Synnaeve and Gabrielle Lee and Georgia Lewis Anderson and Govind Thattai and Graeme Nail and Gregoire Mialon and Guan Pang and Guillem Cucurell and Hailey Nguyen and Hannah Korevaar and Hu Xu and Hugo Touvron and Iliyan Zarov and Imanol Arrieta Ibarra and Isabel Kloumann and Ishan Misra and Ivan Evtimov and Jack Zhang and Jade Copet and Jaewon Lee and Jan Geffert and Jana Vranes and Jason Park and Jay Mahadeokar and Jeet Shah and Jelmer van der Linde and Jennifer Billock and Jenny Hong and Jenya Lee and Jeremy Fu and Jianfeng Chi and Jianyu Huang and Jiawen Liu and Jie Wang and Jiecao Yu and Joanna Bitton and Joe Spisak and Jongsoo Park and Joseph Rocca and Joshua Johnstun and Joshua Saxe and Junteng Jia and Kalyan Vasuden Alwala and Karthik Prasad and Kartikeya Upasani and Kate Plawiak and Ke Li and Kenneth Heafield and Kevin Stone and Khalid El-Arini and Krithika Iyer and Kshitiz Malik and Kuenley Chiu and Kunal Bhalla and Kushal Lakhotia and Lauren Rantala-Yeary and Laurens van der Maaten and Lawrence Chen and Liang Tan and Liz Jenkins and Louis Martin and Lovish Madaan and Lubo Malo and Lukas Blecher and Lukas Landzaat and Luke de Oliveira and Madeline Muzzi and Mahesh Pasupuleti and Mannat Singh and Manohar Paluri and Marcin Kardas and Maria Tsimpoukelli and Mathew Oldham and Mathieu Rita and Maya Pavlova and Melanie Kambadur and Mike Lewis and Min Si and Mitesh Kumar Singh and Mona Hassan and Naman Goyal and Narjes Torabi and Nikolay Bashlykov and Nikolay Bogoychev and Niladri Chatterji and Ning Zhang and Olivier Duchenne and Onur Çelebi and Patrick Alrassy and Pengchuan Zhang and Pengwei Li and Petar Vasic and Peter Weng and Prajjwal Bhargava and Pratik Dubal and Praveen Krishnan and Punit Singh Koura and Puxin Xu and Qing He and Qingxiao Dong and Ragavan Srinivasan and Raj Ganapathy and Ramon Calderer and Ricardo Silveira Cabral and Robert Stojnic and Roberta Raileanu and Rohan Maheswari and Rohit Girdhar and Rohit Patel and Romain Sauvestre and Ronnie Polidoro and Roshan Sumbaly and Ross Taylor and Ruan Silva and Rui Hou and Rui Wang and Saghar Hosseini and Sahana Chennabasappa and Sanjay Singh and Sean Bell and Seohyun Sonia Kim and Sergey Edunov and Shaoliang Nie and Sharan Narang and Sharath Raparthy and Sheng Shen and Shengye Wan and Shruti Bhosale and Shun Zhang and Simon Vandenhende and Soumya Batra and Spencer Whitman and Sten Sootla and Stephane Collot and Suchin Gururangan and Sydney Borodinsky and Tamar Herman and Tara Fowler and Tarek Sheasha and Thomas Georgiou and Thomas Scialom and Tobias Speckbacher and Todor Mihaylov and Tong Xiao and Ujjwal Karn and Vedanuj Goswami and Vibhor Gupta and Vignesh Ramanathan and Viktor Kerkez and Vincent Gonguet and Virginie Do and Vish Vogeti and Vítor Albiero and Vladan Petrovic and Weiwei Chu and Wenhan Xiong and Wenyin Fu and Whitney Meers and Xavier Martinet and Xiaodong Wang and Xiaofang Wang and Xiaoqing Ellen Tan and Xide Xia and Xinfeng Xie and Xuchao Jia and Xuewei Wang and Yaelle Goldschlag and Yashesh Gaur and Yasmine Babaei and Yi Wen and Yiwen Song and Yuchen Zhang and Yue Li and Yuning Mao and Zacharie Delpierre Coudert and Zheng Yan and Zhengxing Chen and Zoe Papakipos and Aaditya Singh and Aayushi Srivastava and Abha Jain and Adam Kelsey and Adam Shajnfeld and Adithya Gangidi and Adolfo Victoria and Ahuva Goldstand and Ajay Menon and Ajay Sharma and Alex Boesenberg and Alexei Baevski and Allie Feinstein and Amanda Kallet and Amit Sangani and Amos Teo and Anam Yunus and Andrei Lupu and Andres Alvarado and Andrew Caples and Andrew Gu and Andrew Ho and Andrew Poulton and Andrew Ryan and Ankit Ramchandani and Annie Dong and Annie Franco and Anuj Goyal and Aparajita Saraf and Arkabandhu Chowdhury and Ashley Gabriel and Ashwin Bharambe and Assaf Eisenman and Azadeh Yazdan and Beau James and Ben Maurer and Benjamin Leonhardi and Bernie Huang and Beth Loyd and Beto De Paola and Bhargavi Paranjape and Bing Liu and Bo Wu and Boyu Ni and Braden Hancock and Bram Wasti and Brandon Spence and Brani Stojkovic and Brian Gamido and Britt Montalvo and Carl Parker and Carly Burton and Catalina Mejia and Ce Liu and Changhan Wang and Changkyu Kim and Chao Zhou and Chester Hu and Ching-Hsiang Chu and Chris Cai and Chris Tindal and Christoph Feichtenhofer and Cynthia Gao and Damon Civin and Dana Beaty and Daniel Kreymer and Daniel Li and David Adkins and David Xu and Davide Testuggine and Delia David and Devi Parikh and Diana Liskovich and Didem Foss and Dingkang Wang and Duc Le and Dustin Holland and Edward Dowling and Eissa Jamil and Elaine Montgomery and Eleonora Presani and Emily Hahn and Emily Wood and Eric-Tuan Le and Erik Brinkman and Esteban Arcaute and Evan Dunbar and Evan Smothers and Fei Sun and Felix Kreuk and Feng Tian and Filippos Kokkinos and Firat Ozgenel and Francesco Caggioni and Frank Kanayet and Frank Seide and Gabriela Medina Florez and Gabriella Schwarz and Gada Badeer and Georgia Swee and Gil Halpern and Grant Herman and Grigory Sizov and Guangyi and Zhang and Guna Lakshminarayanan and Hakan Inan and Hamid Shojanazeri and Han Zou and Hannah Wang and Hanwen Zha and Haroun Habeeb and Harrison Rudolph and Helen Suk and Henry Aspegren and Hunter Goldman and Hongyuan Zhan and Ibrahim Damlaj and Igor Molybog and Igor Tufanov and Ilias Leontiadis and Irina-Elena Veliche and Itai Gat and Jake Weissman and James Geboski and James Kohli and Janice Lam and Japhet Asher and Jean-Baptiste Gaya and Jeff Marcus and Jeff Tang and Jennifer Chan and Jenny Zhen and Jeremy Reizenstein and Jeremy Teboul and Jessica Zhong and Jian Jin and Jingyi Yang and Joe Cummings and Jon Carvill and Jon Shepard and Jonathan McPhie and Jonathan Torres and Josh Ginsburg and Junjie Wang and Kai Wu and Kam Hou U and Karan Saxena and Kartikay Khandelwal and Katayoun Zand and Kathy Matosich and Kaushik Veeraraghavan and Kelly Michelena and Keqian Li and Kiran Jagadeesh and Kun Huang and Kunal Chawla and Kyle Huang and Lailin Chen and Lakshya Garg and Lavender A and Leandro Silva and Lee Bell and Lei Zhang and Liangpeng Guo and Licheng Yu and Liron Moshkovich and Luca Wehrstedt and Madian Khabsa and Manav Avalani and Manish Bhatt and Martynas Mankus and Matan Hasson and Matthew Lennie and Matthias Reso and Maxim Groshev and Maxim Naumov and Maya Lathi and Meghan Keneally and Miao Liu and Michael L. Seltzer and Michal Valko and Michelle Restrepo and Mihir Patel and Mik Vyatskov and Mikayel Samvelyan and Mike Clark and Mike Macey and Mike Wang and Miquel Jubert Hermoso and Mo Metanat and Mohammad Rastegari and Munish Bansal and Nandhini Santhanam and Natascha Parks and Natasha White and Navyata Bawa and Nayan Singhal and Nick Egebo and Nicolas Usunier and Nikhil Mehta and Nikolay Pavlovich Laptev and Ning Dong and Norman Cheng and Oleg Chernoguz and Olivia Hart and Omkar Salpekar and Ozlem Kalinli and Parkin Kent and Parth Parekh and Paul Saab and Pavan Balaji and Pedro Rittner and Philip Bontrager and Pierre Roux and Piotr Dollar and Polina Zvyagina and Prashant Ratanchandani and Pritish Yuvraj and Qian Liang and Rachad Alao and Rachel Rodriguez and Rafi Ayub and Raghotham Murthy and Raghu Nayani and Rahul Mitra and Rangaprabhu Parthasarathy and Raymond Li and Rebekkah Hogan and Robin Battey and Rocky Wang and Russ Howes and Ruty Rinott and Sachin Mehta and Sachin Siby and Sai Jayesh Bondu and Samyak Datta and Sara Chugh and Sara Hunt and Sargun Dhillon and Sasha Sidorov and Satadru Pan and Saurabh Mahajan and Saurabh Verma and Seiji Yamamoto and Sharadh Ramaswamy and Shaun Lindsay and Shaun Lindsay and Sheng Feng and Shenghao Lin and Shengxin Cindy Zha and Shishir Patil and Shiva Shankar and Shuqiang Zhang and Shuqiang Zhang and Sinong Wang and Sneha Agarwal and Soji Sajuyigbe and Soumith Chintala and Stephanie Max and Stephen Chen and Steve Kehoe and Steve Satterfield and Sudarshan Govindaprasad and Sumit Gupta and Summer Deng and Sungmin Cho and Sunny Virk and Suraj Subramanian and Sy Choudhury and Sydney Goldman and Tal Remez and Tamar Glaser and Tamara Best and Thilo Koehler and Thomas Robinson and Tianhe Li and Tianjun Zhang and Tim Matthews and Timothy Chou and Tzook Shaked and Varun Vontimitta and Victoria Ajayi and Victoria Montanez and Vijai Mohan and Vinay Satish Kumar and Vishal Mangla and Vlad Ionescu and Vlad Poenaru and Vlad Tiberiu Mihailescu and Vladimir Ivanov and Wei Li and Wenchen Wang and Wenwen Jiang and Wes Bouaziz and Will Constable and Xiaocheng Tang and Xiaojian Wu and Xiaolan Wang and Xilun Wu and Xinbo Gao and Yaniv Kleinman and Yanjun Chen and Ye Hu and Ye Jia and Ye Qi and Yenda Li and Yilin Zhang and Ying Zhang and Yossi Adi and Youngjin Nam and Yu and Wang and Yu Zhao and Yuchen Hao and Yundi Qian and Yunlu Li and Yuzi He and Zach Rait and Zachary DeVito and Zef Rosnbrick and Zhaoduo Wen and Zhenyu Yang and Zhiwei Zhao and Zhiyu Ma},
      year={2024},
      eprint={2407.21783},
      archivePrefix={arXiv},
      primaryClass={cs.AI},
      url={https://arxiv.org/abs/2407.21783}, 
}

@misc{singh2026openaigpt5card,
      title={OpenAI GPT-5 System Card}, 
      author={Aaditya Singh and Adam Fry and Adam Perelman and Adam Tart and Adi Ganesh and Ahmed El-Kishky and Aidan McLaughlin and Aiden Low and AJ Ostrow and Akhila Ananthram and Akshay Nathan and Alan Luo and Alec Helyar and Aleksander Madry and Aleksandr Efremov and Aleksandra Spyra and Alex Baker-Whitcomb and Alex Beutel and Alex Karpenko and Alex Makelov and Alex Neitz and Alex Wei and Alexandra Barr and Alexandre Kirchmeyer and Alexey Ivanov and Alexi Christakis and Alistair Gillespie and Allison Tam and Ally Bennett and Alvin Wan and Alyssa Huang and Amy McDonald Sandjideh and Amy Yang and Ananya Kumar and Andre Saraiva and Andrea Vallone and Andrei Gheorghe and Andres Garcia Garcia and Andrew Braunstein and Andrew Liu and Andrew Schmidt and Andrey Mereskin and Andrey Mishchenko and Andy Applebaum and Andy Rogerson and Ann Rajan and Annie Wei and Anoop Kotha and Anubha Srivastava and Anushree Agrawal and Arun Vijayvergiya and Ashley Tyra and Ashvin Nair and Avi Nayak and Ben Eggers and Bessie Ji and Beth Hoover and Bill Chen and Blair Chen and Boaz Barak and Borys Minaiev and Botao Hao and Bowen Baker and Brad Lightcap and Brandon McKinzie and Brandon Wang and Brendan Quinn and Brian Fioca and Brian Hsu and Brian Yang and Brian Yu and Brian Zhang and Brittany Brenner and Callie Riggins Zetino and Cameron Raymond and Camillo Lugaresi and Carolina Paz and Cary Hudson and Cedric Whitney and Chak Li and Charles Chen and Charlotte Cole and Chelsea Voss and Chen Ding and Chen Shen and Chengdu Huang and Chris Colby and Chris Hallacy and Chris Koch and Chris Lu and Christina Kaplan and Christina Kim and CJ Minott-Henriques and Cliff Frey and Cody Yu and Coley Czarnecki and Colin Reid and Colin Wei and Cory Decareaux and Cristina Scheau and Cyril Zhang and Cyrus Forbes and Da Tang and Dakota Goldberg and Dan Roberts and Dana Palmie and Daniel Kappler and Daniel Levine and Daniel Wright and Dave Leo and David Lin and David Robinson and Declan Grabb and Derek Chen and Derek Lim and Derek Salama and Dibya Bhattacharjee and Dimitris Tsipras and Dinghua Li and Dingli Yu and DJ Strouse and Drew Williams and Dylan Hunn and Ed Bayes and Edwin Arbus and Ekin Akyurek and Elaine Ya Le and Elana Widmann and Eli Yani and Elizabeth Proehl and Enis Sert and Enoch Cheung and Eri Schwartz and Eric Han and Eric Jiang and Eric Mitchell and Eric Sigler and Eric Wallace and Erik Ritter and Erin Kavanaugh and Evan Mays and Evgenii Nikishin and Fangyuan Li and Felipe Petroski Such and Filipe de Avila Belbute Peres and Filippo Raso and Florent Bekerman and Foivos Tsimpourlas and Fotis Chantzis and Francis Song and Francis Zhang and Gaby Raila and Garrett McGrath and Gary Briggs and Gary Yang and Giambattista Parascandolo and Gildas Chabot and Grace Kim and Grace Zhao and Gregory Valiant and Guillaume Leclerc and Hadi Salman and Hanson Wang and Hao Sheng and Haoming Jiang and Haoyu Wang and Haozhun Jin and Harshit Sikchi and Heather Schmidt and Henry Aspegren and Honglin Chen and Huida Qiu and Hunter Lightman and Ian Covert and Ian Kivlichan and Ian Silber and Ian Sohl and Ibrahim Hammoud and Ignasi Clavera and Ikai Lan and Ilge Akkaya and Ilya Kostrikov and Irina Kofman and Isak Etinger and Ishaan Singal and Jackie Hehir and Jacob Huh and Jacqueline Pan and Jake Wilczynski and Jakub Pachocki and James Lee and James Quinn and Jamie Kiros and Janvi Kalra and Jasmyn Samaroo and Jason Wang and Jason Wolfe and Jay Chen and Jay Wang and Jean Harb and Jeffrey Han and Jeffrey Wang and Jennifer Zhao and Jeremy Chen and Jerene Yang and Jerry Tworek and Jesse Chand and Jessica Landon and Jessica Liang and Ji Lin and Jiancheng Liu and Jianfeng Wang and Jie Tang and Jihan Yin and Joanne Jang and Joel Morris and Joey Flynn and Johannes Ferstad and Johannes Heidecke and John Fishbein and John Hallman and Jonah Grant and Jonathan Chien and Jonathan Gordon and Jongsoo Park and Jordan Liss and Jos Kraaijeveld and Joseph Guay and Joseph Mo and Josh Lawson and Josh McGrath and Joshua Vendrow and Joy Jiao and Julian Lee and Julie Steele and Julie Wang and Junhua Mao and Kai Chen and Kai Hayashi and Kai Xiao and Kamyar Salahi and Kan Wu and Karan Sekhri and Karan Sharma and Karan Singhal and Karen Li and Kenny Nguyen and Keren Gu-Lemberg and Kevin King and Kevin Liu and Kevin Stone and Kevin Yu and Kristen Ying and Kristian Georgiev and Kristie Lim and Kushal Tirumala and Kyle Miller and Lama Ahmad and Larry Lv and Laura Clare and Laurance Fauconnet and Lauren Itow and Lauren Yang and Laurentia Romaniuk and Leah Anise and Lee Byron and Leher Pathak and Leon Maksin and Leyan Lo and Leyton Ho and Li Jing and Liang Wu and Liang Xiong and Lien Mamitsuka and Lin Yang and Lindsay McCallum and Lindsey Held and Liz Bourgeois and Logan Engstrom and Lorenz Kuhn and Louis Feuvrier and Lu Zhang and Lucas Switzer and Lukas Kondraciuk and Lukasz Kaiser and Manas Joglekar and Mandeep Singh and Mandip Shah and Manuka Stratta and Marcus Williams and Mark Chen and Mark Sun and Marselus Cayton and Martin Li and Marvin Zhang and Marwan Aljubeh and Matt Nichols and Matthew Haines and Max Schwarzer and Mayank Gupta and Meghan Shah and Melody Y. Guan and Melody Huang and Meng Dong and Mengqing Wang and Mia Glaese and Micah Carroll and Michael Lampe and Michael Malek and Michael Sharman and Michael Zhang and Michele Wang and Michelle Pokrass and Mihai Florian and Mikhail Pavlov and Miles Wang and Ming Chen and Mingxuan Wang and Minnia Feng and Mo Bavarian and Molly Lin and Moose Abdool and Mostafa Rohaninejad and Nacho Soto and Natalie Staudacher and Natan LaFontaine and Nathan Marwell and Nelson Liu and Nick Preston and Nick Turley and Nicklas Ansman and Nicole Blades and Nikil Pancha and Nikita Mikhaylin and Niko Felix and Nikunj Handa and Nishant Rai and Nitish Keskar and Noam Brown and Ofir Nachum and Oleg Boiko and Oleg Murk and Olivia Watkins and Oona Gleeson and Pamela Mishkin and Patryk Lesiewicz and Paul Baltescu and Pavel Belov and Peter Zhokhov and Philip Pronin and Phillip Guo and Phoebe Thacker and Qi Liu and Qiming Yuan and Qinghua Liu and Rachel Dias and Rachel Puckett and Rahul Arora and Ravi Teja Mullapudi and Raz Gaon and Reah Miyara and Rennie Song and Rishabh Aggarwal and RJ Marsan and Robel Yemiru and Robert Xiong and Rohan Kshirsagar and Rohan Nuttall and Roman Tsiupa and Ronen Eldan and Rose Wang and Roshan James and Roy Ziv and Rui Shu and Ruslan Nigmatullin and Saachi Jain and Saam Talaie and Sam Altman and Sam Arnesen and Sam Toizer and Sam Toyer and Samuel Miserendino and Sandhini Agarwal and Sarah Yoo and Savannah Heon and Scott Ethersmith and Sean Grove and Sean Taylor and Sebastien Bubeck and Sever Banesiu and Shaokyi Amdo and Shengjia Zhao and Sherwin Wu and Shibani Santurkar and Shiyu Zhao and Shraman Ray Chaudhuri and Shreyas Krishnaswamy and Shuaiqi and Xia and Shuyang Cheng and Shyamal Anadkat and Simón Posada Fishman and Simon Tobin and Siyuan Fu and Somay Jain and Song Mei and Sonya Egoian and Spencer Kim and Spug Golden and SQ Mah and Steph Lin and Stephen Imm and Steve Sharpe and Steve Yadlowsky and Sulman Choudhry and Sungwon Eum and Suvansh Sanjeev and Tabarak Khan and Tal Stramer and Tao Wang and Tao Xin and Tarun Gogineni and Taya Christianson and Ted Sanders and Tejal Patwardhan and Thomas Degry and Thomas Shadwell and Tianfu Fu and Tianshi Gao and Timur Garipov and Tina Sriskandarajah and Toki Sherbakov and Tomek Korbak and Tomer Kaftan and Tomo Hiratsuka and Tongzhou Wang and Tony Song and Tony Zhao and Troy Peterson and Val Kharitonov and Victoria Chernova and Vineet Kosaraju and Vishal Kuo and Vitchyr Pong and Vivek Verma and Vlad Petrov and Wanning Jiang and Weixing Zhang and Wenda Zhou and Wenlei Xie and Wenting Zhan and Wes McCabe and Will DePue and Will Ellsworth and Wulfie Bain and Wyatt Thompson and Xiangning Chen and Xiangyu Qi and Xin Xiang and Xinwei Shi and Yann Dubois and Yaodong Yu and Yara Khakbaz and Yifan Wu and Yilei Qian and Yin Tat Lee and Yinbo Chen and Yizhen Zhang and Yizhong Xiong and Yonglong Tian and Young Cha and Yu Bai and Yu Yang and Yuan Yuan and Yuanzhi Li and Yufeng Zhang and Yuguang Yang and Yujia Jin and Yun Jiang and Yunyun Wang and Yushi Wang and Yutian Liu and Zach Stubenvoll and Zehao Dou and Zheng Wu and Zhigang Wang},
      year={2026},
      eprint={2601.03267},
      archivePrefix={arXiv},
      primaryClass={cs.CL},
      url={https://arxiv.org/abs/2601.03267}, 
}
\clearpage

\appendix
\section{Detailed Related Work}
\label{app:relatedWork}
\subsection{LLM-Based Planning and Agents}
LLMs have demonstrated strong capabilities in reasoning and structured decision-making tasks \citep{wei2022chain, yao2023react}. Prior work explores several strategies for improving LLM planning performance, including chain-of-thought reasoning \citep{wei2022chain}, deliberative strategies such as Tree-of-Thought \citep{yao2023tree}, and iterative self-reflection \citep{shinn2024reflexion}. 

More recently, multi-agent systems have been proposed to decompose complex tasks into specialized reasoning components. Frameworks such as Atlas \citep{choi2026atlas} MetaGPT \citep{hong2024metagpt} and CAMEL \citep{li2023camel} demonstrate that agent-based collaboration can improve reasoning efficiency for complex tasks. Despite these advances, LLMs still struggle to reliably satisfy complex constraints in long-horizon planning \citep{valmeekam2023large}. While some research translates LLM outputs into formal Planning Domain Definition Language (PDDL) for execution \citep{liu2023llmpddl}, scaling these to the preference-driven realities of human travel remains a challenge.

\subsection{Travel Planning Benchmarks and Systems}
Travel planning has recently emerged as an important domain for evaluating LLM planning capabilities. Benchmarks such as \textbf{TravelPlanner} \citep{xie2024travelplanner} and \textbf{TripCraft} \citep{chaudhuri2025tripcraft} formulate itinerary generation as a multi-constraint reasoning task involving fine-grained spatio-temporal constraints. Retrieval-based approaches like TP-RAG \citep{ni2025tprag} use retrieval-augmented frameworks to guide point-of-interest (POI) sequences. However, these primarily rely on structural databases or past trajectories and do not explicitly incorporate the high-density qualitative information found in real-world user reviews.

\paragraph{LLM-as-a-Judge Evaluation.}
Evaluating automated travel generation has traditionally relied on deterministic constraint-checking alongside quantitative spatial and temporal metrics \citep{chaudhuri2025tripcraft, xie2024travelplanner}. While effective for measuring structural feasibility and routing efficiency, these quantitative metrics struggle to capture the subjective, qualitative success of an itinerary; such as ambiance, service quality, and safety. Recently, the \textit{LLM-as-a-Judge} paradigm has emerged as a robust alternative for evaluating complex, open-ended generative tasks that lack a single ground-truth reference \citep{zheng2023judging}. We extend this paradigm to the domain of travel planning by proposing a review-grounded evaluation metric. By equipping the LLM judge with distilled user reviews, our framework bridges the gap between TripCraft's mathematically rigid quantitative scores and human-centric experiential quality.

\subsection{Hybrid and Formal Planning Approaches}
Another line of research combines LLM reasoning with classical planning algorithms or symbolic solvers to improve constraint satisfaction. Formal verification approaches translate natural language queries into symbolic constraints and solve them using Satisfiability Modulo Theories (SMT) \citep{BT18, hao2025formalplanning}. While these methods excel at improving structural constraint feasibility, they abstract away the semantic nuance of travel entities. This makes it difficult to directly optimize for qualitative aspects like ``coziness'' or safety risks without manually hardcoding exhaustive ontologies. Our work bridges this gap by using agents to handle semantic nuances while a deterministic backend ensures feasibility.

\subsection{Review-Aware Recommendation Systems}
The review distillation process used in our framework represents the latest evolution of Review-Aware Recommender Systems (RARS). Early techniques focused on extracting latent features from text to mitigate the sparsity of explicit numerical ratings. 
\begin{itemize}
    \item \textbf{Collaborative Filtering with Reviews:} Early models such as DeepCoNN \citep{zheng2017joint} used dual convolutional neural networks to model user preferences and item properties from review text simultaneously.
    \item \textbf{Aspect-Based Sentiment Analysis (ABSA):} Techniques like those proposed in \citep{pontiki2014semeval} focused on identifying specific entity aspects (e.g., ``Food Quality'') and the sentiment attached to them.
\end{itemize}

The selection of travel entities shares underlying principles with review-driven recommendation. Recent work such as \textit{RevBrowse} \citep{wei2025learning} demonstrates that LLMs can mimic human shopping behavior by structuring reviews to highlight salient attributes, effectively managing the constrained context window of LLMs. While traditional POI recommendation uses reviews to mitigate the sparsity of explicit ratings \citep{10.1145/3178876.3186070}, our work is the first to port these distilled review summaries structured as \textit{Pros} and \textit{Cons}—directly into the constraint-aware entity selection loops of an autonomous planning framework. These are not used merely for ranking, but as immediate reasoning primitives that our planning agents use to satisfy complex user personas.

\subsection{Embodied Agents and Open-World Planning}
\label{appendix:embodied}
While \system{} operates in the digital-semantic domain of travel planning, the modularity of our architecture draws inspiration from embodied agent frameworks. Systems such as \textbf{SayCan} \citep{ahn2022can} use LLMs to generate high-level plans that are grounded in physical robotic affordances via a value function. Similarly, \textbf{Voyager} \citep{wang2023voyager} demonstrates autonomous skill discovery and long-term planning in open-ended environments such as Minecraft via an automated curriculum. 

Although travel planning does not involve physical motor control, the challenge of grounding ``reasoning'' (e.g., selecting a hotel) into ``executable constraints'' (e.g., staying under a \$500 budget) shares the same structural foundation. By moving beyond monolithic prompting to specialized modules, our framework adopts the ``divide-and-conquer'' strategy successful in open-world embodied planning.

\subsection{Formal Planning and PDDL Grounding}
\label{appendix:formal}
To bridge the gap between natural language and deterministic execution, recent research has explored translating LLM outputs into \textbf{Planning Domain Definition Language (PDDL)} \citep{liu2023llmpddl, guan2023leveraging}. These frameworks use formal solvers to verify that an LLM’s plan is physically or logically possible within a defined world model. 

In the travel domain, however, defining a complete PDDL world model is difficult due to the constant flux of real-world data (e.g., changing flight prices or restaurant hours). \system{} avoids the rigid overhead of PDDL by using a hybrid backend. This allows us to maintain the semantic richness of the LLM while enforcing hard constraints (budget, time, distance) through a deterministic algorithmic controller, ensuring reliability without sacrificing the qualitative nuance found in reviews.

\paragraph{Our Approach.}
In contrast to prior work, \system{} uses agentic task decomposition as the architectural mechanism that enables review-grounded planning. Incorporating large volumes of real-world user reviews is essential for capturing qualitative aspects of travel, but directly injecting such high-density textual data into a monolithic LLM leads to severe reasoning overload. By decomposing planning into specialized domain modules (accommodations, transportation, etc.), our framework creates localized context windows that allow review-derived information—distilled as \textit{Pros} and \textit{Cons} to be processed efficiently within each agent. This design prevents reasoning overload while enabling the system to prioritize entities based on distilled user experiences. The resulting framework combines the semantic flexibility of LLMs with the reliability of deterministic constraint enforcement, thereby unifying structural feasibility and user preference modeling within a single coherent system.

\newpage

\begin{table*}[t]
\centering
\scriptsize
\setlength{\tabcolsep}{1pt} 
% \resizebox{\textwidth}{!}{%
\begin{tabular}{lc cccccc cccccc}
\toprule
\multirow{2}{*}{} & \textbf{Dur} 
& \multicolumn{6}{c}{\textbf{Constraint Satisfaction (\%)}} 
& \multicolumn{6}{c}{\textbf{Temporal+Structural}} \\
\cmidrule(lr){3-8} \cmidrule(l){9-14}
& (day) & Del & CPR$_\mu$  &CPR$_M$ & $ HCPR_\mu$ & HCPR$_M$ & FPR & $T_m$ & $T_a$ & $\tilde{T}_a$ & $S_s$ & $S_p$ & $S_o$ \\
\midrule
\multicolumn{14}{l}{\textit{\textbf{Baseline: Monolithic Prompting (Original TripCraft~\cite{chaudhuri2025tripcraft})}}} \\
\midrule
\multirow{3}{*}{\rotatebox{90}{GPT5}} & 3& 100 & 81.54 & 1.74 & 94.49 & 55.23 & 1.45 & .79 & .17 & -- & .81 & .49 & .80 \\
               & 5& 100 & 67.47 & 0 & 92.92 & 23.15 & 0 & .84 & .22 & -- & .83 & .49 & .95 \\
               & 7& 99.70 & 57.28 & 0 & 90.12 & 13.85 & 0 & .86 & .22 & -- & .87 & .50 & .97 \\
\hline
\multirow{3}{*}{\rotatebox{90}{Phi4}} & 3& 92.60 & 47.69 & .0 & .0 & .0 & .0 & .24 & .22 & -- & .60 & .53 & .66 \\
      & 5& 99.56 & 43.86 & .0 & .0 & .0 & .0 & .53 & .13 & -- & .83 & .53 & .92 \\
      & 7& 97.79 & 37.22 & .0 & .0 & .0 & .0 & .51 & .14 & -- & .83 & .53 & .96 \\
\hline
\multirow{3}{*}{\rotatebox{90}{Qwen2.5}} & 3& 99.56 & 70.39 & .43 & 3.26 & 2.60 & .0 & .58 & .15 & -- & .869 & .51 & .79 \\
         & 5& 99.56 & 52.30 & .0 & .0 & .0 & .0 & .59 & .09 & -- & .742 & .51 & .92 \\
         & 7& 99.13 & 39.29 & .0 & .0 & .0 & .0 & .57 & .01 & -- & .734 & .52 & .96\\

\midrule
\multicolumn{14}{l}{\textit{\textbf{Baseline: Agentic Framework (Original ATLAS~\cite{choi2026atlas})}}} \\
\midrule
\multirow{3}{*}{\rotatebox{90}{Qwen2.5}} & 3& 100 & 53.22 & 0 & 1.27 & 0.34 & 0 & 0.28 & 0.07 & -- & 0.70 & 0.50 & 0.60 \\
         & 5& 100 & 37.1 & 0 & 0 & 0 & 0 & 0.19 & 0.03 & -- & 0.56 & 0.46 & 0.85 \\
         & 7& 100 & 34.14 & 0 & 0 & 0 & 0 & 0.20 & 0.05 & -- & 0.61 & 0.48 & 0.91\\         
\midrule
\multicolumn{14}{l}{\textit{\textbf{\system{} (Ours): LLM Scheduler}}} \\
\midrule
\multirow{3}{*}{\rotatebox{90}{GPT5}}& 3& 100 & 94.39 & 64.83 & \textbf{91.02} & \textbf{83.72} & 52.03 & \textbf{.87} & .13 & .36 & .86 & .51 & .80 \\
      & 5& 100 & 90.83 & 44.44 & 87.23 & 71.60 & 33.33 & .78 & .25 & .71 & .95 & .51 & .93 \\
      & 7& 100 & 80.69 & 7.53 & 87.72 & 76.81 & 5.42 & .74 & .25 & .73 & \textbf{\underline{.95}} & .50 & .96 \\
\hline
\multirow{3}{*}{\rotatebox{90}{Phi4}}& 3& 94.48 & 83.88 & 12.50 & 89.96 & 71.22 & 10.17 & .83 & .21 & .58 & .88 & \textbf{\underline{.52}} & .66 \\
      & 5& 89.20 & 75.74 & 1.54 & \textbf{88.86} & 62.04 & 1.54 & .80 & .20 & .55 & .93 & .51 & .89 \\
      & 7& 92.77 & 71.10 & 2.11 & 86.88 & 49.40 & 1.20 & .80 & .18 & .53 & \textbf{\underline{.95}} & \textbf{\underline{.52 }} & .95 \\
\hline
\multirow{3}{*}{\rotatebox{90}{Qwen2.5}} & 3& 99.42 & 77.65 & .00 & 80.93 & 66.28 & .00 & .50 & .10 & .29 & .87 & \textbf{\underline{.52}} & .62 \\
         & 5& 96.91 & 67.25 & .00 & 82.74 & 66.36 & .00 & .46 & .15 & .43 & .95 & \textbf{\underline{.52 }}& .88 \vspace{1pt}\\
         & 7& 97.59 & 65.57 & .00 & 81.51 & 68.37 & .00 & .49 & .15 & .42 & .93 & \textbf{\underline{.52 }}& .94 \vspace{1pt}\\
\hline
\multirow{3}{*}{\rotatebox{90}{Llama3.1}}& 3& 95.63 & 77.06 & .29 & 82.90 & 72.67 & .29 & .62 & .15 & .42 & .89 & \textbf{\underline{.52}} & .67 \vspace{1pt}\\
         & 5& 85.19 & 62.65 & .0 & 72.34 & 57.72 & .0 & .59 & .12 & .33 & .93 & \textbf{\underline{.52}} & .89 \vspace{1pt}\\
         & 7& 99.70 & 69.67 & .0 & 83.25 & 68.07 & .0 & .56 & .11 & .31 & .92 & \textbf{\underline{.52}} & .95 \vspace{1pt}\\
\hline
\multirow{3}{*}{\rotatebox{90}{DS-R1}}& 3& 78.78 & 61.05 & 2.91 & 68.39 & 61.92 & 2.33 & .73 & .22 & .61 & \textbf{\underline{.90}} & \textbf{\underline{.52}} & .66 \\
         & 5& 53.40 & 35.50 & .0 & 45.15 & 34.88 & .0 & .72 & .20 & .58 & .93 & .51 & .88 \\
         & 7& 37.95 & 23.89 & .0 & 32.26 & 29.22 & .0 & .72 & .17 & .50 & .92 & .52 & .94 \\
\hline
\multirow{3}{*}{\rotatebox{90}{M-Nemo}} & 3& 98.84 & 77.06 & 1.16 & 85.61 & 73.55 & 1.16 & .56 & .15 & .43 & .89 & \textbf{\underline{.52}} & .65 \\
         & 5& 90.12 & 66.27 & .0 & 76.83 & 61.42 & .0 & .53 & .13 & .36 & .94 & \textbf{\underline{.52 }}& .89 \\
         & 7& 90.96 & 65.93 & .0 & 77.30 & 62.05 & .0 & .52 & .12 & .34 & .92 & \textbf{\underline{.52}} & .95 \\
\midrule
\multicolumn{14}{l}{\textit{\textbf{\system{} (Ours): Deterministic Algorithmic Scheduler}}} \\
\midrule
\multirow{3}{*}{\rotatebox{90}{GPT5}}& 3& 100 & 98.92 & 89.83 & 90.9 & 83.43 & \textbf{74.42} & .81 & \textbf{\underline{.26}} & \textbf{.76} & .76 & .51 & \textbf{.83} \\
      & 5& 100 & 98.61 & 86.11 & 88.06 & \textbf{73.15} & 62.04 & .78 & .25 & .71 & .95 & .51 & .93 \\
      & 7& 100 & 98.80 & 88.25 & \textbf{88.34} & \textbf{78.61} & \textbf{68.67} & .82 & \textbf{\underline{.26}} & \textbf{\underline{.75}} & \textbf{\underline{.95}} & .50 & \textbf{\underline{.97}} \\
\hline
\multirow{3}{*}{\rotatebox{90}{Phi4}}& 3& 98.26 & 98.76 & 86.05 & 84.77 & 59.88 & 59.88 & .82 & .24 & .69 & .88 & .51 & .80 \\
      & 5& 95.99 & \textbf{99.94} & \textbf{95.37} & 85.70 & 65.12 & \textbf{65.12} & .79 & \textbf{\underline{.26}} & \textbf{\underline{.73}} & .93 & .50 & \textbf{\underline{.94}} \\
      & 7& 98.80 & 99.51 & 93.98 & 85.21 & 66.27 & 65.96 & .82 & \textbf{\underline{.26}} & \textbf{\underline{.75}} & .93 & .51 & \textbf{\underline{.97}} \\
\hline
\multirow{3}{*}{\rotatebox{90}{Qwen2.5}}& 3& 100 & \textbf{99.30} & 93.02 & 81.55 & 68.31 & 62.79 & .82 & .24 & .70 & .87 & \textbf{\underline{.52}} & .73 \\
         & 5& 99.69 & 98.70 & 89.81 & 86.64 & 69.14 & 62.04 & \textbf{.81} & .24 & .69 & .94 & .51 & .91 \vspace{1pt}\\
         & 7& 95.48 & 99.02 & 86.45 & 85.45 & 60.54 & 60.54 & .82 & .25 & .72 & .93 & \textbf{\underline{.52}} & .96 \vspace{1pt}\\
\hline
\multirow{3}{*}{\rotatebox{90}{Llama3.1}}& 3& 95.64 & 95.64 & \textbf{95.64} & 79.46 & 66.28 & 66.28 & .82 & \textbf{\underline{.26}} & .75 & .88 & .51 & .80 \vspace{1pt}\\
         & 5& 85.19 & 85.19 & 85.19 & 71.75 & 56.79 & 56.79 & .79 & \textbf{\underline{.26}} & \textbf{\underline{.73}} & .94 & .50 & \textbf{\underline{.94}} \vspace{1pt}\\
         & 7& 100 & \textbf{99.52} & \textbf{95.18} & 82.38 & 67.17 & 63.25 & .82 & \textbf{\underline{.26}} & \textbf{\underline{.75}} & .92 & .50 & \textbf{\underline{.97 }}\vspace{1pt}\\
\hline
\multirow{3}{*}{\rotatebox{90}{DS-R1}}& 3& 79.07 & 78.92 & 77.62 & 66.91 & 59.59 & 58.14 & .83 & .25 & .75 & \textbf{\underline{.90}} & .51 & .81 \\
         & 5& 56.17 & 56.17 & 56.17 & 47.40 & 37.04 & 37.04 & .80 & \textbf{\underline{.26}} & \textbf{\underline{.73}} & .93 & .50 & \textbf{\underline{.94 }}\\
         & 7& 41.87 & 41.66 & 39.76 & 34.74 & 32.53 & 30.72 & \textbf{.83} & .25 & .73 & .93 & .51 & \textbf{\underline{.97}} \\
\hline
\multirow{3}{*}{\rotatebox{90}{M-Nemo}}& 3& 98.84 & 98.28 & 93.60 & 84.87 & 71.80 & 67.73 & .81 & .25 & .72 & .88 & .51 & .81 \\
         & 5& 91.36 & 91.30 & 90.74 & 75.65 & 57.72 & 57.10 & .80 & .25 & .72 & .93 & .51 & \textbf{\underline{.94 }}\\
         & 7& 93.67 & 93.37 & 90.66 & 78.16 & 62.35 & 59.94 & .82 & .25 & .73 & .92 & .51 & \textbf{\underline{.97}} \\
\bottomrule
\end{tabular}%
\caption{Performance of \system{} on constraint satisfaction, temporal and structural metrics. We compare the LLM-based scheduler with the deterministic algorithmic scheduler against the monolithic, agentic baselines. $T_m$= Temporal meal score, $T_a$= Temporal attraction score,  $\tilde{T}_a$=normalized temporal attraction score (introduced in ~\system{}), $S_s$=spatial score, $S_p$=persona alignment score, and $S_o$=ordering score. DS-R1=DepSeek-R1, M-Nemo=Mistral-Nemo. Bold values represents the best value across all the models. \underline{Underlined} values signifies of equal values  }
\label{tab:tripcraft_complete_results}
\end{table*}

\section{Implementation Details and Hyperparameters}
\label{app:hyperparams}
Our framework evaluates both proprietary and open-weight models. The proprietary baseline (GPT-5) was accessed via the official OpenAI developer API to ensure standard, reproducible closed-source evaluation. The open-weight models (including Llama 3.1, DeepSeek-R1, Phi-4, Mistral-Nemo, and Qwen 2.5) were deployed locally using the Hugging Face Transformers library in PyTorch. Local models were loaded in \texttt{bfloat16} precision with automated device mapping. 

To ensure strict adherence to JSON output formats and deterministic constraint satisfaction, generation hyperparameters were optimized across our model suite:
\begin{itemize}
    \item \textbf{General Search Space:} Across all models, we constrained the sampling space using $top\_p = 0.9$ and generated up to $2000$ new tokens. 
    \item \textbf{Open-Weight Hyperparameters:} For the locally deployed models, we utilized a baseline sampling temperature of $T = 0.6$. To prevent degenerative loops during JSON generation, we applied a repetition penalty of $1.05$. Furthermore, specific system prompts enforcing structural compliance and ChatML templates were applied via the tokenizer's \texttt{apply\_chat\_template} function to ensure valid, parseable outputs.
\end{itemize}

\begin{table}[htbp]
\centering
\footnotesize
\resizebox{\columnwidth}{!}{%
\begin{tabular}{lr}
\toprule
\textbf{Constant Description} & \textbf{Value (minutes)} \\
\midrule
\multicolumn{2}{c}{\textit{Global Constraints}} \\
\midrule
Travel Buffer ($\Delta_{transit}$) & $30$ \\
Minimum Meal Gap ($\gamma$) & $240$ \\
Stay Duration (Check-in/out) & $30$ \\
\midrule
\multicolumn{2}{c}{\textit{Required Durations}} \\
\midrule
Breakfast & $50$ \\
Lunch & $60$ \\
Dinner & $75$ \\
Attraction (Base) & $120$ \\
Overnight Stay & $480$ \\
\midrule
\multicolumn{2}{c}{\textit{Permissible Windows (Start Times)}} \\
\midrule
Breakfast & $[8 \times 60,\; 10 \times 60 + 30]$ \\
Lunch & $[12 \times 60,\; 15 \times 60 + 51]$ \\
Dinner & $[18 \times 60 + 30,\; 22 \times 60 + 30]$ \\
Attractions & $[9 \times 60,\; 19 \times 60]$ \\
\bottomrule
\end{tabular}%
}
\caption{Temporal scheduling parameters. Values for durations and windows are adopted from the TripCraft framework \cite{chaudhuri2025tripcraft}.}
\label{tab:temporal_constants}
\end{table}

\section{Example Travel Query Formulations}
\label{app:query_example}

To illustrate the problem formulation defined in Section 3.1, we present three complete examples of travel queries of varying lengths (3-day, 5-day, and 7-day) from our evaluation framework. We demonstrate how the natural language text is mapped to the formal tuple structure $q = (c_s, C_d, W, k, B, C_{local}, \pi)$ via our query extraction module.

\subsection{Example 1: 3-Day Trip (Single Destination)}

\paragraph{Natural Language Query:}
\textit{``Plan a 3-day trip for 8 people from Montgomery to Washington from November 8th to November 10th, 2024, with a budget of \$8,400. Accommodations should include a private room and allow smoking. The itinerary should feature visits to sights and landmarks as well as concerts and shows.''}

\paragraph{Assigned Traveler Persona:}
\textit{``Traveler Type: Laidback Traveler; Purpose of Travel: Nature; Spending Preference: Economical Traveler; Location Preference: Mountains.''}

\paragraph{Structured Tuple Extraction:}
\begin{itemize}
    \item \textbf{Source City ($c_s$):} Montgomery
    \item \textbf{Destination Cities ($C_d$):} \{\text{Washington}\}
    \item \textbf{Temporal Window ($W$):} 3 days (2024-11-08 to 2024-11-10)
    \item \textbf{Group Size ($k$):} 8
    \item \textbf{Total Budget ($B$):} \$8,400.00
    \item \textbf{Local Constraints ($C_{local}$):} 
    \begin{itemize}
        \item \texttt{house\_rule}: smoking
        \item \texttt{room\_type}: private room
        \item \texttt{attraction}: [Sights \& Landmarks, Concerts \& Shows]
    \end{itemize}
    \item \textbf{Persona ($\pi$):} \{Laidback, Nature, Economical, Mountains\}
\end{itemize}

\subsection{Example 2: 5-Day Trip (Multi-City)}

\paragraph{Natural Language Query:}
\textit{``Design a 5-day travel itinerary for 3 people starting from Denver and visiting 2 cities in Iowa between November 8th and November 12th, 2024. The budget is \$5,400. The trip should avoid flights and include Mexican cuisine options as well as visits to nature and parks attractions.''}

\paragraph{Assigned Traveler Persona:}
\textit{``Traveler Type: Laidback Traveler; Purpose of Travel: Nature; Spending Preference: Economical Traveler; Location Preference: Mountains.''}

\paragraph{Structured Tuple Extraction:}
\begin{itemize}
    \item \textbf{Source City ($c_s$):} Denver
    \item \textbf{Destination State ($C_d$):} \{\text{Iowa (2 cities)}\}
    \item \textbf{Temporal Window ($W$):} 5 days (2024-11-08 to 2024-11-12)
    \item \textbf{Group Size ($k$):} 3
    \item \textbf{Total Budget ($B$):} \$5,400.00
    \item \textbf{Local Constraints ($C_{local}$):} 
    \begin{itemize}
        \item \texttt{cuisine}: Mexican
        \item \texttt{transportation}: no flight
        \item \texttt{attraction}: Nature \& Parks
    \end{itemize}
    \item \textbf{Persona ($\pi$):} \{Laidback, Nature, Economical, Mountains\}
\end{itemize}

\subsection{Example 3: 7-Day Trip (Multi-City)}

\paragraph{Natural Language Query:}
\textit{``Design a 7-day travel itinerary for 1 person starting from San Antonio and visiting 3 cities in Tennessee from November 3rd to November 9th, 2024. The budget is \$7,150. The trip should avoid flights as a mode of transportation and include events related to sports and arts \& theatre. Accommodations must allow visitors.''}

\paragraph{Assigned Traveler Persona:}
\textit{``Traveler Type: Adventure Seeker; Purpose of Travel: Relaxation; Spending Preference: Economical Traveler; Location Preference: Cities.''}

\paragraph{Structured Tuple Extraction:}
\begin{itemize}
    \item \textbf{Source City ($c_s$):} San Antonio
    \item \textbf{Destination State ($C_d$):} \{\text{Tennessee (3 cities)}\}
    \item \textbf{Temporal Window ($W$):} 7 days (2024-11-03 to 2024-11-09)
    \item \textbf{Group Size ($k$):} 1
    \item \textbf{Total Budget ($B$):} \$7,150.00
    \item \textbf{Local Constraints ($C_{local}$):} 
    \begin{itemize}
        \item \texttt{house\_rule}: visitors
        \item \texttt{transportation}: no flight
        \item \texttt{event}: [Sports, Arts \& Theatre]
    \end{itemize}
    \item \textbf{Persona ($\pi$):} \{Adventure Seeker, Relaxation, Economical, Cities\}
\end{itemize}

\section{Table of Notations}
\label{app:notations}

Table~\ref{tab:notations} provides a comprehensive summary of the formal mathematical notations and variables used throughout the \system{} framework, categorized by their role in the pipeline.

\begin{table*}[t]
\centering
\small
\renewcommand{\arraystretch}{1.2}
\begin{tabular}{lp{13cm}}
\toprule
\textbf{Notation} & \textbf{Description} \\
\midrule
\multicolumn{2}{c}{\textit{Query \& Input Parameters}} \\
\midrule
$q$ & Formalized travel query tuple. \\
$c_s$ & Source (origin) city. \\
$C_d$ & Set of destination cities. \\
$W$ & Temporal window (travel dates and duration). \\
$k$ & Group size (number of travelers). \\
$B$ & Total trip budget constraint. \\
$C_{local}$ & Local constraints extracted from the query (e.g., house rules, cuisine). \\
$\pi$ & Multi-dimensional traveler persona (e.g., Budget, Nature, Laidback). \\
\midrule
\multicolumn{2}{c}{\textit{Database \& Constraints}} \\
\midrule
$\mathcal{DB}$ & Complete travel database. \\
$\mathcal{DB}_{acc}$ & Subset of database containing accommodations. \\
$\mathcal{DB}_{trans}$ & Subset of database containing transportation options. \\
$\mathcal{DB}_{rest}$ & Subset of database containing restaurants. \\
$\mathcal{DB}_{attr}$ & Subset of database containing tourist attractions. \\
$\mathcal{DB}_{event}$ & Subset of database containing time-bound events. \\
$\mathcal{R}_{temp}$ & Set of temporal rules and scheduling constants. \\
$\Delta_{transit}$ & Transit buffer required between consecutive activities. \\
$\gamma$ & Minimum gap required between meals. \\
\midrule
\multicolumn{2}{c}{\textit{Pipeline Variables \& Outputs}} \\
\midrule
$A_{acc}$ & Selected accommodation entity for a specific city. \\
$A_{trans}$ & Selected global transportation plan. \\
$R_{rank}$ & List of ranked restaurant candidates. \\
$T_{rank}$ & List of ranked tourist attraction candidates. \\
$E_{opt}$ & List of optional time-bound event candidates. \\
$S$ & Fixed timeline skeleton (transport and accommodation boundaries). \\
$I_{unsched}$ & Unscheduled itinerary containing all selected entities. \\
$I$ & Final, fully scheduled spatio-temporal itinerary. \\
$e_i$ & A single selected travel entity (POI) within the itinerary. \\
$t_i^{start}, t_i^{end}$ & Start and end times for a scheduled entity $e_i$. \\
$N$ & Total number of scheduled items in the final itinerary. \\
\bottomrule
\end{tabular}
\caption{Summary of notations used in the \system{} formulation and algorithms.}
\label{tab:notations}
\end{table*}

\section{Refining TripCraft Temporal Evaluators}
\label{app:temporal_metric}

This section provides the implementation details and mathematical justification for our refinements to the temporal evaluation scripts provided by the original TripCraft benchmark.

\subsection{Temporal Overlap Resolution}
In the original commonsense constraint evaluation script, POI visits could logically overlap with inter-city departure times without being flagged by the evaluator. To clarify, this fix operates as a patch to the constraint verification logic, rather than a standalone metric. 

We corrected this by enforcing strict arrival and departure boundary checks against the transportation strings, thereby correctly failing physically impossible schedules.

\subsection{Original Formulation and Limitation of Temporal Attraction Score}
The original Temporal Attraction Score is designed to evaluate whether the time allocated to attractions is realistic. It is formulated as a joint probabilistic model combining a continuous distribution for visit duration and a discrete Poisson distribution for the daily attraction count:

$$f_{X,N}(d_i, n) = f_{X|N}(d_i \mid n) \cdot P(N=n)$$

The conditional duration likelihood is modeled using a Gaussian distribution:

$$f_{X|N}(d_i \mid n) = \exp\left(-\frac{(d_i - \mu_d^i)^2}{2\sigma_d^2}\right)$$

where $d_i$ is the actual duration spent, $\mu_d^i$ is the expected duration, and $\sigma_d$ captures variability. The number of attractions visited per day is modeled via a Poisson distribution:

$$P(N=n) = \frac{\lambda^n e^{-\lambda}}{n!}$$

where $\lambda$ represents the expected number of attractions for a given traveler persona. The final score is computed by averaging this joint likelihood over all attractions:

$$\bar{T}_{\text{attrac}} = \frac{1}{n} \sum_{i=1}^n \exp\left(-\frac{(d_i - \mu_d^i)^2}{2\sigma_d^2}\right) \cdot \frac{\lambda^n e^{-\lambda}}{n!}$$

\textbf{Limitation:} Because the Poisson term is a probability mass function, its maximum value is strictly less than $1$. For a typical laid-back traveler ($\lambda \approx 1.11$), the maximum Poisson probability is approximately $0.36$. Consequently, even if a generated itinerary is perfectly realistic (where the Gaussian term evaluates to $1$), the overall Temporal Attraction Score is artificially capped around $0.35$. Numerically low values falsely appear as poor performance when they actually represent near-optimal itineraries.

\subsection{Normalized Improvement}
To resolve this artificial cap and scale the metric to a standard intuitive range of $[0, 1]$, we normalize the Poisson term by its maximum theoretical value for a given persona:

$$\tilde{P}(N=n) = \frac{P(N=n)}{\max_k P(N=k)}$$

This normalization preserves the relative behavioral likelihoods of different attraction counts while allowing a perfect, highly realistic itinerary to achieve a score of $1.0$, enabling clearer differentiation between good and excellent plans.

\section{Case Study: Generative vs. Deterministic Scheduling}
\label{app:scheduling_failures}

To illustrate the fundamental limitations of pure LLM-based temporal scheduling, we present a comparative case study of a 3-day itinerary from Green Bay to Atlanta. Both pipelines utilized identical intermediate agent outputs (the same selected flights, accommodations, and ranked restaurants); however, they diverged in how the final temporal sequence was constructed.

\subsection{Failures of the Generative LLM Scheduler}
When tasked with assigning specific timestamps and transit buffers, the generative LLM produced severe chronological and logical errors:

\begin{itemize}
    \item \textbf{Day 1 (Duration Failure):} The LLM scheduled the user to stay at the ``Downtown Beach Room'' from \texttt{10:42 to 10:42} and again from \texttt{12:00 to 12:00}. It failed to comprehend that a stay requires a positive temporal duration.
    \item \textbf{Day 1 (Transit Overlap):} The LLM scheduled a visit to Max's Coal Oven Pizzeria from \texttt{12:00 to 13:00}, perfectly overlapping with the hallucinated hotel stay and leaving zero transit buffer between locations.
    \item \textbf{Day 3 (Circadian Violation):} Despite being prompted with standard daylight constraints, the LLM scheduled lunch at Ray's in the City from \texttt{03:30 to 04:30} in the morning.
\end{itemize}

\subsection{Success of the Deterministic Algorithmic Scheduler}
Using the exact same entity selections, the deterministic algorithmic backend successfully resolved the spatio-temporal puzzle without hallucinations:

\begin{itemize}
    \item \textbf{Day 1 (Corrected Durations and Transit):} The scheduler logically processed the flight arrival (09:42), added a standard transit buffer, and scheduled the initial hotel drop-off from \texttt{10:12 to 10:42}. It then successfully allocated the required transit buffer (41 meters to nearest transit) before scheduling the Fox Theatre visit from \texttt{11:12 to 13:24}.
    \item \textbf{Day 2 \& 3 (Strict Sequencing):} The algorithm maintained strict chronological order and enforced appropriate meal windows, successfully scheduling a normal daytime lunch at The Americano from \texttt{14:40 to 15:40} and dinner at Nikolai's Roof from \texttt{20:45 to 22:00}.
\end{itemize}

This case study confirms that while multi-agent frameworks are exceptional at semantic entity selection, the actual assignment of temporal windows is best handled by programmatic, deterministic execution to guarantee feasibility.

\section{Expected Output Format}
\label{app:json_format}

To ensure standardized evaluation by the algorithmic scheduler and scoring functions, \system{} outputs the final itinerary in a strictly formatted JSON structure. The \texttt{point\_of\_interest\_list} is a semi-structured string parsed via regex to enforce temporal boundaries and distance metrics. 

Below are examples of the schema across varying trip durations.

\subsection{3-Day Trip Example (Flight)}

\begin{Verbatim}[fontsize=\small, breaklines=true, frame=single]
{
  "days": [
    {
      "day": 1,
      "current_city": "from Sitka to Seattle",
      "transportation": "Flight Number: F2879280, from Sitka to Seattle, Departure Time: 06:50, Arrival Time: 10:04",
      "breakfast": "-",
      "lunch": "Wild Ginger, Seattle",
      "dinner": "Noi Thai Cuisine, Seattle",
      "attraction": "The Museum of Flight, Seattle;",
      "accommodation": "Peaceful Classic Seattle Neighborhood, Seattle",
      "event": "-",
      "point_of_interest_list": "Peaceful Classic Seattle Neighborhood, stay from 10:34 to 11:04, nearest transit: Roosevelt Way NE & NE 80th St, 315.01m away; The Museum of Flight, visit from 11:34 to 14:01, nearest transit: East Marginal Way S & S 94th Pl, 95.97m away; Wild Ginger, visit from 14:40 to 15:40, nearest transit: 3rd Ave & Pike St, 13.86m away; Noi Thai Cuisine, visit from 20:45 to 22:00, nearest transit: 2nd Ave & Seneca St, 181.65m away; Peaceful Classic Seattle Neighborhood, stay from 22:00 to 08:00, nearest transit: Roosevelt Way NE & NE 80th St, 315.01m away"
    },
    {
      "day": 2,
      "current_city": "Seattle",
      "transportation": "-",
      "breakfast": "Lola, Seattle",
      "lunch": "Barolo Ristorante, Seattle",
      "dinner": "All Water Seafood & Oyster Bar, Seattle",
      "attraction": "Museum of Pop Culture, Seattle; Chihuly Garden and Glass, Seattle;",
      "accommodation": "Peaceful Classic Seattle Neighborhood, Seattle",
      "event": "-",
      "point_of_interest_list": "Peaceful Classic Seattle Neighborhood, stay from 08:00 to 08:30, nearest transit: Roosevelt Way NE & NE 80th St, 315.01m away; Lola, visit from 09:30 to 10:20, nearest transit: Virginia St & 4th Ave, 26.1m away; Museum of Pop Culture, visit from 10:50 to 13:17, nearest transit: 5th Ave N & Broad St, 53.97m away; Barolo Ristorante, visit from 14:40 to 15:40, nearest transit: Westlake And 7th, 37.77m away; Chihuly Garden and Glass, visit from 16:10 to 18:37, nearest transit: Seattle Center, 78.76m away; All Water Seafood & Oyster Bar, visit from 20:45 to 22:00, nearest transit: 1st Ave & Spring St, 19.43m away; Peaceful Classic Seattle Neighborhood, stay from 22:00 to 08:00, nearest transit: Roosevelt Way NE & NE 80th St, 315.01m away"
    },
    {
      "day": 3,
      "current_city": "from Seattle to Sitka",
      "transportation": "Flight Number: F2337064, from Seattle to Sitka, Departure Time: 17:12, Arrival Time: 18:31",
      "breakfast": "Bacco Cafe, Seattle",
      "lunch": "Dough Zone Dumpling House - Seattle International District, Seattle",
      "dinner": "-",
      "attraction": "Klondike Gold Rush National Historical Park, Seattle;",
      "accommodation": "-",
      "event": "-",
      "point_of_interest_list": "Peaceful Classic Seattle Neighborhood, stay from 08:00 to 08:30, nearest transit: Roosevelt Way NE & NE 80th St, 315.01m away; Bacco Cafe, visit from 09:30 to 10:20, nearest transit: 2nd Ave & Stewart St, 107.26m away; Klondike Gold Rush National Historical Park, visit from 10:50 to 14:02, nearest transit: Occidental Mall, 113.08m away; Dough Zone Dumpling House - Seattle International District, visit from 14:40 to 15:40, nearest transit: 5th Ave S & S Weller St, 26.54m away"
    }
  ]
}
\end{Verbatim}

\subsection{5-Day Trip Example (Self-Driving)}

\begin{Verbatim}[fontsize=\small, breaklines=true, frame=single]
{
  "days": [
    {
      "day": 1,
      "current_city": "from Peoria to Memphis",
      "transportation": "Self-Driving from Peoria to Memphis, Duration: 495 mins, Departure Time: 06:00, Arrival Time: 14:15",
      "breakfast": "-",
      "lunch": "-",
      "dinner": "Brother Juniper's, Memphis",
      "attraction": "Memphis Zoo, Memphis;",
      "accommodation": "Cozy 1 Bedroom Guest House, Memphis",
      "event": "A Beautiful Noise (Touring), Memphis",
      "point_of_interest_list": "Cozy 1 Bedroom Guest House, stay from 14:45 to 15:15, nearest transit: ECHLES@CARNES, 246.18m away; Memphis Zoo, visit from 15:45 to 18:52, nearest transit: N PARKWAY@WEST DR, 8.68m away; Brother Juniper's, visit from 20:45 to 22:00, nearest transit: SOUTHERN@ELLSWORTH, 199.7m away; Cozy 1 Bedroom Guest House, stay from 22:00 to 08:00, nearest transit: ECHLES@CARNES, 246.18m away"
    },
    {
      "day": 2,
      "current_city": "Memphis",
      "transportation": "-",
      "breakfast": "The Brass Door Irish Pub, Memphis",
      "lunch": "Leonard's Pit Barbecue, Memphis",
      "dinner": "Memphis BBQ Grill, Memphis",
      "attraction": "Memphis Riverboats, Memphis; Backbeat Tours, Memphis;",
      "accommodation": "Cozy 1 Bedroom Guest House, Memphis",
      "event": "Rod Wave - Last Lap Tour, Memphis",
      "point_of_interest_list": "Cozy 1 Bedroom Guest House, stay from 08:00 to 08:30, nearest transit: ECHLES@CARNES, 246.18m away; The Brass Door Irish Pub, visit from 09:30 to 10:20, nearest transit: THIRD@COURT, 76.39m away; Memphis Riverboats, visit from 10:50 to 13:47, nearest transit: RIVERSIDE DR @ COURT AVE, 213.38m away; Leonard's Pit Barbecue, visit from 14:40 to 15:40, nearest transit: MTMORIAH RD@MENDENHALL, 214.04m away; Backbeat Tours, visit from 16:10 to 19:07, nearest transit: THIRD@PEABODYPL, 78.56m away; Memphis BBQ Grill, visit from 20:45 to 22:00; Cozy 1 Bedroom Guest House, stay from 22:00 to 05:00, nearest transit: ECHLES@CARNES, 246.18m away"
    },
    {
      "day": 3,
      "current_city": "from Memphis to Chattanooga",
      "transportation": "Self-Driving from Memphis to Chattanooga, Duration: 373 mins, Departure Time: 06:00, Arrival Time: 12:13",
      "breakfast": "-",
      "lunch": "Bridgeman's Chophouse, Chattanooga",
      "dinner": "Alleia, Chattanooga",
      "attraction": "Lookout Mountain, Chattanooga;",
      "accommodation": "Cozy Stylish Studio, Chattanooga",
      "event": "All-Ages Micro Wrestling at the Microtorium of Pigeon Forge, Chattanooga",
      "point_of_interest_list": "Cozy 1 Bedroom Guest House, stay from 05:00 to 05:30; Cozy Stylish Studio, stay from 12:43 to 13:13, nearest transit: BONNOAK, 4223.21m away; Bridgeman's Chophouse, visit from 14:40 to 15:40, nearest transit: BROAD + READ HOUSE, 27.65m away; Lookout Mountain, visit from 16:10 to 19:22, nearest transit: 55TH + ALABAMA, 1650.28m away; Alleia, visit from 20:45 to 22:00, nearest transit: Main & Market Outbound, 49.84m away; Cozy Stylish Studio, stay from 22:00 to 08:00, nearest transit: BONNOAK, 4223.21m away"
    },
    {
      "day": 4,
      "current_city": "Chattanooga",
      "transportation": "-",
      "breakfast": "Hennen's, Chattanooga",
      "lunch": "Easy Bistro & Bar, Chattanooga",
      "dinner": "St. John's Restaurant, Chattanooga",
      "attraction": "Chattanooga Zoo, Chattanooga; Chattanooga Ducks, Chattanooga;",
      "accommodation": "Cozy Stylish Studio, Chattanooga",
      "event": "Tennessee Volunteers Volleyball vs. Texas A&M Volleyball, Chattanooga",
      "point_of_interest_list": "Cozy Stylish Studio, stay from 08:00 to 08:30, nearest transit: BONNOAK, 4223.21m away; Hennen's, visit from 09:30 to 10:20, nearest transit: SHUTTLE PARK NORTH - INTERNAL, 112.36m away; Chattanooga Ducks, visit from 10:50 to 13:47, nearest transit: SHUTTLE PARK NORTH - INTERNAL, 413.96m away; Easy Bistro & Bar, visit from 14:40 to 15:40, nearest transit: BRO AQ 1, 56.72m away; Chattanooga Zoo, visit from 16:10 to 19:17, nearest transit: Holtzclaw & 5th-1, 123.88m away; St. John's Restaurant, visit from 20:45 to 22:00, nearest transit: Market & King1, 25.22m away; Cozy Stylish Studio, stay from 22:00 to 08:00, nearest transit: BONNOAK, 4223.21m away"
    },
    {
      "day": 5,
      "current_city": "from Chattanooga to Peoria",
      "transportation": "Self-Driving from Chattanooga to Peoria, Duration: 633 mins, Departure Time: 16:00",
      "breakfast": "Zaya 1943 Korean Steakhouse, Chattanooga",
      "lunch": "The Purple Daisy Picnic Cafe, Chattanooga",
      "dinner": "-",
      "attraction": "Tennessee Riverwalk, Chattanooga;",
      "accommodation": "-",
      "event": "-",
      "point_of_interest_list": "Cozy Stylish Studio, stay from 08:00 to 08:30, nearest transit: BONNOAK, 4223.21m away; Zaya 1943 Korean Steakhouse, visit from 09:30 to 10:20, nearest transit: MANUFACTURERS RD + CHEROKEE BLVD, 152.9m away; Tennessee Riverwalk, visit from 10:50 to 13:17, nearest transit: Amnicol a& River Terminal-1, 178.82m away; The Purple Daisy Picnic Cafe, visit from 14:20 to 15:20, nearest transit: St. Elmo & 40th1, 36.0m away"
    }
  ]
}
\end{Verbatim}

\subsection{7-Day Trip Example (Taxi)}
\begin{Verbatim}[fontsize=\small, breaklines=true, frame=single]
{
  "days": [
    {
      "day": 1,
      "current_city": "from Cleveland to Nashville",
      "transportation": "Taxi from Cleveland to Nashville, Duration: 583 mins, Departure Time: 06:00, Arrival Time: 15:43",
      "breakfast": "-",
      "lunch": "-",
      "dinner": "The Catbird Seat, Nashville",
      "attraction": "Cheekwood, Nashville;",
      "accommodation": "Fancy Stay Walk to Broadway Park Free Let\u2019s Roll!, Nashville",
      "event": "Mondo Cozmo with Special Guest Jane Leo, Nashville",
      "point_of_interest_list": "Fancy Stay Walk to Broadway Park Free Let\u2019s Roll!, stay from 16:13 to 16:43, nearest transit: KOREAN VETS BLVD & 6TH AVE WB, 123.82m away; Cheekwood, visit from 17:13 to 20:10, nearest transit: HWY 70 S & BROOK HOLLOW RD EB, 833.18m away; The Catbird Seat, visit from 20:45 to 22:00, nearest transit: BROADWAY AVE & 17TH AVE S EB, 189.47m away; Fancy Stay Walk to Broadway Park Free Let\u2019s Roll!, stay from 22:00 to 08:00, nearest transit: KOREAN VETS BLVD & 6TH AVE WB, 123.82m away"
    },
    {
      "day": 2,
      "current_city": "Nashville",
      "transportation": "-",
      "breakfast": "Luogo, Nashville",
      "lunch": "Bourbon Steak by Michael Mina, a Nashville Steakhouse, Nashville",
      "dinner": "The Chef And I, Nashville",
      "attraction": "Gaylord Opryland Garden Conservatory, Nashville; Centennial Park, Nashville;",
      "accommodation": "Fancy Stay Walk to Broadway Park Free Let\u2019s Roll!, Nashville",
      "event": "Get the Led Out, Nashville",
      "point_of_interest_list": "Fancy Stay Walk to Broadway Park Free Let\u2019s Roll!, stay from 08:00 to 08:30, nearest transit: KOREAN VETS BLVD & 6TH AVE WB, 123.82m away; Luogo, visit from 09:00 to 09:50, nearest transit: 12TH AVE & LAUREL ST NB, 86.33m away; Centennial Park, visit from 10:20 to 14:02, nearest transit: WEST END AVE & 27TH AVE S WB, 441.64m away; Bourbon Steak by Michael Mina, a Nashville Steakhouse, visit from 14:40 to 15:40, nearest transit: 8TH AVE S & DEMONBREUN ST NB, 99.41m away; Gaylord Opryland Garden Conservatory, visit from 16:10 to 20:07, nearest transit: OPRY MILLS DRIVE & WARDROBE BLDNG, 471.91m away; The Chef And I, visit from 20:45 to 22:00, nearest transit: 21ST AVE & BROADWAY AVE SB, 157.63m away; Fancy Stay Walk to Broadway Park Free Let\u2019s Roll!, stay from 22:00 to 05:00, nearest transit: KOREAN VETS BLVD & 6TH AVE WB, 123.82m away"
    },
    {
      "day": 3,
      "current_city": "from Nashville to Memphis",
      "transportation": "Taxi from Nashville to Memphis, Duration: 235 mins, Departure Time: 06:00, Arrival Time: 09:55",
      "breakfast": "-",
      "lunch": "Brother Juniper's, Memphis",
      "dinner": "The Brass Door Irish Pub, Memphis",
      "attraction": "Graceland, Memphis;",
      "accommodation": "Private room/bath in shared home, Memphis",
      "event": "David Nihill: Shelf Help Tour, Memphis",
      "point_of_interest_list": "Fancy Stay Walk to Broadway Park Free Let\u2019s Roll!, stay from 05:00 to 05:30; Private room/bath in shared home, stay from 10:25 to 10:55, nearest transit: CENTRAL AVE@LAFAYETTE ST, 400.89m away; Graceland, visit from 11:25 to 13:52, nearest transit: ELVIS PRESLEY BLVD@DOLAN DR, 199.9m away; Brother Juniper's, visit from 14:40 to 15:40, nearest transit: SOUTHERN@ELLSWORTH, 199.7m away; The Brass Door Irish Pub, visit from 20:45 to 22:00, nearest transit: THIRD@COURT, 76.39m away; Private room/bath in shared home, stay from 22:00 to 08:00, nearest transit: CENTRAL AVE@LAFAYETTE ST, 400.89m away"
    },
    {
      "day": 4,
      "current_city": "Memphis",
      "transportation": "-",
      "breakfast": "Memphis BBQ Grill, Memphis",
      "lunch": "Central BBQ, Memphis",
      "dinner": "Flight Restaurant and Wine Bar, Memphis",
      "attraction": "Sun Studio, Memphis; Stax Museum of American Soul Music, Memphis;",
      "accommodation": "Private room/bath in shared home, Memphis",
      "event": "-",
      "point_of_interest_list": "Private room/bath in shared home, stay from 08:00 to 08:30, nearest transit: CENTRAL AVE@LAFAYETTE ST, 400.89m away; Memphis BBQ Grill, visit from 09:30 to 10:20; Sun Studio, visit from 10:50 to 13:17, nearest transit: UNION AVE@MARSHALL AVE, 90.18m away; Central BBQ, visit from 14:40 to 15:40, nearest transit: SUMMER AVE@SANDRIDGE, 20.99m away; Stax Museum of American Soul Music, visit from 16:10 to 18:37, nearest transit: MCLEMORE@COLLEGE, 36.06m away; Flight Restaurant and Wine Bar, visit from 20:45 to 22:00, nearest transit: UNION-MAIN/TRO SB, 103.36m away; Private room/bath in shared home, stay from 22:00 to 05:00, nearest transit: CENTRAL AVE@LAFAYETTE ST, 400.89m away"
    },
    {
      "day": 5,
      "current_city": "from Memphis to Chattanooga",
      "transportation": "Taxi from Memphis to Chattanooga, Duration: 373 mins, Departure Time: 06:00, Arrival Time: 12:13",
      "breakfast": "-",
      "lunch": "Bridgeman's Chophouse, Chattanooga",
      "dinner": "Alleia, Chattanooga",
      "attraction": "Lookout Mountain, Chattanooga;",
      "accommodation": "New Town House in Historic District of Chattanooga, Chattanooga",
      "event": "Disney On Ice presents Into the Magic, Chattanooga",
      "point_of_interest_list": "Private room/bath in shared home, stay from 05:00 to 05:30; New Town House in Historic District of Chattanooga, stay from 12:43 to 13:13, nearest transit: S MARKET + 17TH, 37.73m away; Bridgeman's Chophouse, visit from 14:40 to 15:40, nearest transit: BROAD + READ HOUSE, 27.65m away; Lookout Mountain, visit from 16:10 to 19:22, nearest transit: 55TH + ALABAMA, 1650.28m away; Alleia, visit from 20:45 to 22:00, nearest transit: Main & Market Outbound, 49.84m away; New Town House in Historic District of Chattanooga, stay from 22:00 to 08:00, nearest transit: S MARKET + 17TH, 37.73m away"
    },
    {
      "day": 6,
      "current_city": "Chattanooga",
      "transportation": "-",
      "breakfast": "Hennen's, Chattanooga",
      "lunch": "Easy Bistro & Bar, Chattanooga",
      "dinner": "St. John's Restaurant, Chattanooga",
      "attraction": "The Lookout Mountain Incline Railway, Chattanooga; Tennessee Valley Railroad Museum, Chattanooga;",
      "accommodation": "New Town House in Historic District of Chattanooga, Chattanooga",
      "event": "-",
      "point_of_interest_list": "New Town House in Historic District of Chattanooga, stay from 08:00 to 08:30, nearest transit: S MARKET + 17TH, 37.73m away; Hennen's, visit from 09:30 to 10:20, nearest transit: SHUTTLE PARK NORTH - INTERNAL, 112.36m away; The Lookout Mountain Incline Railway, visit from 10:50 to 13:17, nearest transit: TENN AVE + INCLINE, 63.48m away; Easy Bistro & Bar, visit from 14:40 to 15:40, nearest transit: BRO AQ 1, 56.72m away; Tennessee Valley Railroad Museum, visit from 16:10 to 18:37, nearest transit: Bonny Oaks & Redlands Dr-1, 1191.41m away; St. John's Restaurant, visit from 20:45 to 22:00, nearest transit: Market & King1, 25.22m away; New Town House in Historic District of Chattanooga, stay from 22:00 to 08:00, nearest transit: S MARKET + 17TH, 37.73m away"
    },
    {
      "day": 7,
      "current_city": "from Chattanooga to Cleveland",
      "transportation": "Taxi from Chattanooga to Cleveland, Duration: 679 mins, Departure Time: 16:00",
      "breakfast": "Zaya 1943 Korean Steakhouse, Chattanooga",
      "lunch": "The Purple Daisy Picnic Cafe, Chattanooga",
      "dinner": "-",
      "attraction": "Chattanooga Zoo, Chattanooga;",
      "accommodation": "-",
      "event": "-",
      "point_of_interest_list": "New Town House in Historic District of Chattanooga, stay from 08:00 to 08:30, nearest transit: S MARKET + 17TH, 37.73m away; Zaya 1943 Korean Steakhouse, visit from 09:30 to 10:20, nearest transit: MANUFACTURERS RD + CHEROKEE BLVD, 152.9m away; Chattanooga Zoo, visit from 10:50 to 13:57, nearest transit: Holtzclaw & 5th-1, 123.88m away; The Purple Daisy Picnic Cafe, visit from 14:20 to 15:20, nearest transit: St. Elmo & 40th1, 36.0m away"
    }
  ]
}
\end{Verbatim}

\section{Qualitative Case Studies}
\label{app:case_studies}

To analyze the impact of review-grounded reasoning, we present qualitative case studies comparing itineraries generated with and without review integration. These examples illustrate how structured review ``Pros'' and ``Cons'' enable the framework to align with nuanced personas while avoiding experiential failure modes. Unlike quantitative metrics, these cases provide interpretable evidence of how review-aware planning improves attraction selection, dining, and overall coherence, reflecting the preferences captured by our LLM-as-a-Judge evaluator.

\subsection{Case Study 1: Cultural and Artistic Persona Alignment}

\paragraph{Traveler Persona:}
\textit{Traveler Type: Cultural Explorer; Spending Preference: Moderate; Purpose of Travel: Arts and Local Experiences.}

\paragraph{LLM Judge Preference Rationale:}
The evaluator strongly preferred the review-grounded itinerary for its superior alignment with the traveler's artistic focus and improved dining authenticity.

\begin{table}[!t]
\centering
\small
\caption{Entity replacements after review integration for the Everett itinerary.}
\begin{tabular}{p{0.3\linewidth} p{0.3\linewidth} p{0.3\linewidth}}
\toprule
\textbf{Without Review} & \textbf{With Review} & \textbf{Observed Improvement} \\
\midrule
Flying Heritage \& Combat Armor Museum & Schack Art Center & Stronger artistic and cultural alignment \\
\hline
Papa Everett's Pizza & Abbondanza Ristorante & More authentic dining experience \\
\hline
Beautiful House A on Gibson Rd. & Large Room with Private Bathroom & Improved accommodation comfort \\
\bottomrule
\end{tabular}
\end{table}

\paragraph{Review Evidence:}
\textbf{Schack Art Center (Added):} ``Beautiful local artwork, creative workshops, and an immersive artistic atmosphere.''\\
\textbf{Flying Heritage \& Combat Armor Museum (Removed):} ``Primarily focused on military vehicles and combat exhibits.''

\paragraph{Analysis:}
The review-aware planner replaced operationally generic entities with culturally immersive alternatives. Substituting a military museum for an art center demonstrates that review-grounded reasoning enables semantically aligned selection beyond simple category matching.

\subsection{Case Study 2: Scenic and Nature-Oriented Travel Planning}

\paragraph{Traveler Persona:}
\textit{Traveler Type: Relaxed Traveler; Purpose of Travel: Nature and Scenic Exploration; Spending Preference: Luxury Traveler.}

\paragraph{LLM Judge Preference Rationale:}
The review-grounded itinerary was favored for prioritizing peaceful coastal immersion over crowded urban landmarks.

\begin{table}[!t]
\centering
\small
\caption{Entity replacements after review integration for the San Francisco itinerary.}
\begin{tabular}{p{0.3\linewidth} p{0.3\linewidth} p{0.3\linewidth}}
\toprule
\textbf{Without Review} & \textbf{With Review} & \textbf{Observed Improvement} \\
\midrule
Golden Gate Park & Lands End & Improved scenic and coastal immersion \\
\hline
City View Restaurant & Eight AM & Higher experiential dining quality \\
\hline
Generic bakery stop & Highly rated local dining & More authentic local experience \\
\bottomrule
\end{tabular}
\end{table}

\paragraph{Review Evidence:}
\textbf{Lands End (Added):} ``Amazing coastal trails, breathtaking views, and peaceful scenic exploration.''\\
\textbf{Golden Gate Park (Removed):} ``Large urban park with crowded areas during peak hours.''

\paragraph{Analysis:}
The framework prioritized scenic immersion and authentic nature experiences over crowded urban parks, proving it captures nuanced experiential suitability beyond basic popularity metrics.

\subsection{Case Study 3: Multi-City Experiential Optimization}

\paragraph{Traveler Persona:}
\textit{Traveler Type: Laidback Traveler; Purpose of Travel: Nature; Spending Preference: Economical Traveler; Location Preference: Mountains.}

\paragraph{LLM Judge Preference Rationale:}
The evaluator noted a substantial improvement in Risk Avoidance, as the review-aware variant successfully filtered out venues with known hygiene and service issues.

\begin{table}[!t]
\centering
\small
\caption{Representative entity replacements in the multi-city itinerary.}
\begin{tabular}{p{0.3\linewidth} p{0.3\linewidth} p{0.3\linewidth}}
\toprule
\textbf{Without Review} & \textbf{With Review} & \textbf{Observed Improvement} \\
\midrule
Amarillo Zoo & Richard Daniel Baker Peace Park & More peaceful and reflective experience \\
\hline
Abuelo's Mexican Restaurant & Phoenicia Specialty Foods & Improved food quality and atmosphere \\
\hline
The Plaza Restaurant & Hugo's & Better dining consistency and service \\
\hline
El Manantial & Pappadeaux Seafood Kitchen & Higher experiential dining quality \\
\hline
El Bracero Mexican Grill & Goode Company Seafood & Reduced negative food-quality risks \\
\bottomrule
\end{tabular}
\end{table}

\paragraph{Examples of Avoided Experiential Risks:}
\textbf{Amarillo Zoo (Removed):} ``Boring experience, dirty enclosures, unhelpful staff...''\\
\textbf{Richard Daniel Baker Peace Park (Added):} ``Peaceful escape promoting kindness, reflection, and meaningful experiences...''\\
\textbf{Abuelo's Mexican Restaurant (Removed):} ``Food is bland and unremarkable. Service was poor...''\\
\textbf{Pappadeaux Seafood Kitchen (Added):} ``Excellent food quality, flavorful dishes, pleasant atmosphere...''

\paragraph{Analysis:}
The planner consistently replaced entities associated with overcrowding, poor hygiene, and bad service with venues offering peaceful environments and authentic dining. This multi-city improvement proves the framework optimizes semantic experiential coherence rather than relying on rigid database retrieval.

\subsection{Summary of Qualitative Findings}

Across all case studies, review-grounded reasoning consistently improved persona alignment, dining authenticity, accommodation comfort, and risk avoidance. These qualitative observations strongly support the empirical score increases captured by our LLM-as-a-Judge experiential evaluation framework.

\section{Query Parsing Prompt}
\label{app:query_prompt}

Before agent planning begins, the system extracts structured constraints from the
natural language user query. This step converts the raw query into explicit
planning parameters such as number of travelers, available budget, and any
local constraints.

The extraction module uses a constrained prompt to ensure that only explicitly
stated information is captured and that the resulting JSON strictly matches the
internal schema used by the planning agents.

\begin{Verbatim}[fontsize=\small, breaklines=true, frame=single]
You are a STRICT information extraction system.

Extract ONLY the following fields from the travel query.
Do NOT guess.
Do NOT infer.
If a value is NOT explicitly stated OR defined by the interpretation rules, return null.

Return ONLY valid JSON.
No explanation.
No markdown.

FIELDS:
- people_number (integer)
- budget (number or null)
- local_constraint:
  - house rule (string or null)
  - cuisine (string or null)
  - room type (string or null)
  - transportation (string or null)
  - event (string or null)
  - attraction (string or null)

INTERPRETATION RULES

People count rules:
- Words like "solo", "single", "me", "myself", "alone" → people_number = 1
- Phrases like "two people", "we are two", "couple" → people_number = 2
- Phrases like "group of X", "party of X" → people_number = X

Budget rules:
- "$1900", "$1,900", "budget is 1900", "budget set at $1900" → budget = 1900
- If no numeric budget exists → budget = null

Constraint rules:
- Extract constraints only if they are explicitly mentioned in the query.
- Constraints may refer to accommodation rules, cuisine preferences,
  room types, transportation restrictions, preferred event categories,
  or attraction types.
- If the query says "no specific constraints" → all constraints = null.
- Do NOT invent preferences.

Example Input:
"Plan a 3-day trip for one person from St. Petersburg to Rockford. The budget is $1700."

Example Output:
{
  "people_number": 1,
  "budget": 1700,
  "local_constraint": {
    "house rule": null,
    "cuisine": null,
    "room type": null,
    "transportation": null,
    "event": null,
    "attraction": null
  }
}
\end{Verbatim}

\section{Agent Prompts}
\label{app:prompts}

This section details the specific system prompts used to instruct the various domain agents and the global planner within the \system{} framework.

\subsection{Accommodation Agent Prompt}
\begin{Verbatim}[fontsize=\small, breaklines=true, frame=single]
You are the ACCOMMODATION SELECTION AGENT.
Task: Select EXACTLY ONE accommodation from {{ACCOMMODATION_REF}} that best fits the traveler.

Constraints:
1. Choose the CHEAPEST valid option for the required trip nights.
2. Output the EXACT object (no hallucinated values).
3. STRICTLY enforce House Rules (e.g., pets, smoking, parties, children) based on the Persona.
4. STRICTLY enforce Room Type rules from {{LOCALCONSTRAINTS_JSON}}. 
5. Do NOT extract budget limitations from the persona; rely only on the global constraints.
If ANY house rule or room type constraint is violated, the accommodation MUST be rejected.

Inputs:
- Accommodations: {{ACCOMMODATION_REF}}
- Persona: {{PERSONA_JSON}}
- Local Constraints: {{LOCALCONSTRAINTS_JSON}}

Output strictly in JSON format:
{
  "hotel": { ... exact hotel object from accommodation_ref ... }
}
\end{Verbatim}

\subsection{Transportation Agent Prompt}
\begin{Verbatim}[fontsize=\small, breaklines=true, frame=single]
You are the TRANSPORT PLANNING AGENT.
Task: Select exactly ONE valid `mode_strategy` (flight, taxi, or self-driving) and assign exactly one transport leg per travel day (odd-numbered days: 1, 3, 5...) from {{TRANSPORT_REF_JSON}}.

Mode Selection Pipeline (Strict Order):
1. Candidate Set: Intersect base modes with {{ALLOWED_MODES}}.
2. Local Constraints: STRICTLY enforce {{LOCALCONSTRAINTS_JSON}} (e.g., "no flight", "taxi required"). Remove violating modes.
3. Cost Filter: Total cost across ALL legs must be <= {{TRANSPORT_CAPS_JSON}}. 
   - Note: Flights are per person ({{PEOPLE}}). Taxi/Self-driving are per vehicle.
4. Tie-Breaker: Prefer the candidate with the lower total travel duration.
5. Priority: If still tied, prefer Flight > Taxi > Self-driving, dynamically adjusting if a mode is forbidden.

Timing Rules (Hard Constraints):
- Flights: Any reasonable timing.
- Taxi/Self-driving (> 12 hours): Day 1 arrival MUST be ~19:30. Last day departure MUST be ~15:30-16:00.

Inputs:
- Trip Length: {{DAYS}} days | Travel Days: {{TRAVEL_DAYS}} | People: {{PEOPLE}}
- Persona: {{PERSONA_JSON}} (Guidance only, does not override hard rules)
- References: {{TRANSPORT_REF_JSON}}

Output strictly in JSON format:
{
  "mode_strategy": "flight | taxi | self-driving",
  "legs": [
    {
      "day": <int>,
      "from": "<city>", "to": "<city>",
      "mode": "flight | taxi | self-driving",
      "details": { ... EXACT object from transport_ref ... },
      "departure_time": "HH:MM | null",
      "arrival_time": "HH:MM | null"
    }
  ]
}
\end{Verbatim}

\subsection{Meals Agent Prompt}
\begin{Verbatim}[fontsize=\small, breaklines=true, frame=single]
You are the MEAL SELECTION AGENT.
Task: ORDER at least 12 restaurants from {{RESTAURANT_CARDS_JSON}} by how well they match the user's persona and local constraints.

Ranking Priority & Logic:
1. Top Results: The FIRST 3-4 restaurants MUST satisfy all explicit food-related {{LOCAL_CONSTRAINTS_JSON}} whenever possible.
2. Cuisine Match: Treat cuisines in persona/constraints as POSITIVE, SOFT preferences. Do NOT exclude other cuisines.
3. Persona Signals: Prefer higher persona_alignment and persona_utility.
4. Quality & Risk: Prefer higher aggregate_rating, restaurant_quality, and aspect scores (food, service, ambience). Penalize high wait_risk and hygiene_risk.
5. Budget: Prefer lower avg_cost when quality and persona fit are tied.

Strict Rules:
1. Output ONLY exact restaurant names from the provided list (no hallucinations).
2. Do NOT repeat restaurant names.
3. You MUST return AT LEAST 12 restaurants (if fewer exist, return all).
4. Cuisines are NEVER a hard filter; they only influence ranking.
5. Do NOT calculate total cost, people count, or meal counts.
6. Do NOT extract budget limitations from the persona.

Inputs:
- Meals Budget Cap: {{MEALS_CAP}}
- Restaurants: {{RESTAURANT_CARDS_JSON}}
- Persona: {{PERSONA_JSON}}
- Local Constraints: {{LOCAL_CONSTRAINTS_JSON}}

Output strictly in JSON format:
{
  "restaurants_ranked": [
      "Restaurant Name 1",
      "Restaurant Name 2",
      ...
  ]
}
\end{Verbatim}

\subsection{Attraction Agent Prompt}
\begin{Verbatim}[fontsize=\small, breaklines=true, frame=single]
You are the ATTRACTION SELECTION AGENT.
Task: Select and ORDER 8-12 attractions from {{ATTRACTIONS_JSON}} based on persona match and local constraints. (If fewer than 8 exist, select ALL of them).

Ranking Priority & Logic:
1. Top Results: The FIRST 3-4 attractions SHOULD satisfy explicit attraction-type preferences from {{LOCAL_CONSTRAINTS_JSON}} when possible.
2. Category Match: Treat categories in persona/constraints as POSITIVE, SOFT preferences. Do NOT strictly exclude other attraction types unless they violate safety or physical constraints.
3. Persona Signals: Prefer higher persona_alignment and persona_utility. Align with specific persona themes (e.g., Luxury -> iconic; Adventure -> outdoor; Cultural -> museums; Family -> kid-friendly).
4. Quality & Risk: Prefer higher attraction_quality and aspect signals (experience, nature, culture). Penalize high crowd_risk and safety_risk.
5. Duration: Consider visit_duration suitability (avoid extremely long activities if not reasonable).

Strict Rules:
1. Output ONLY exact attraction names from the provided list (no hallucinations).
2. Do NOT use any external knowledge beyond the provided data.
3. This is a RELATIVE ranking: Earlier items = better match, Later items = weaker match.

Inputs:
- Attractions: {{ATTRACTIONS_JSON}}
- Persona: {{PERSONA_JSON}}
- Local Constraints: {{LOCAL_CONSTRAINTS_JSON}}

Output strictly in JSON format:
{
  "attractions_ranked": [
      "Attraction Name 1",
      "Attraction Name 2",
      "Attraction Name 3",
      ...
  ]
}
\end{Verbatim}

\subsection{Events Agent Prompt}
\begin{Verbatim}[fontsize=\small, breaklines=true, frame=single]
You are the EVENT SELECTION AGENT.
Task: Select AT MOST ONE event per DATE from {{EVENT_CARDS_JSON}}. Events are date-bound and OPTIONAL (returning null is a valid and acceptable action).

Selection Priority & Logic:
1. Category Match: Treat event categories mentioned in the persona and local constraints as POSITIVE, SOFT preferences. Prefer events matching these categories.
2. Conflict Resolution: If multiple preferred events exist on the same date, pick the ONE best match.
3. Soft Filtering: Do NOT automatically exclude events from other categories solely because they aren't explicitly requested. 

Strict Rules:
1. Output ONLY exact event names from the provided list (no hallucinations).
2. Assign a maximum of ONE event per date.
3. If no suitable events are available on a date, or if no preferences strongly match, return `null` for that date.

Inputs:
- Events: {{EVENT_CARDS_JSON}}
- Persona: {{PERSONA_JSON}}
- Local Constraints: {{LOCAL_CONSTRAINTS_JSON}}

Output strictly in JSON format:
{
  "events_by_date": {
    "YYYY-MM-DD": "Event Name",
    "YYYY-MM-DD": null
  }
}
\end{Verbatim}

\subsection{Itinerary Skeleton Filler (Planner) Prompt}
\begin{Verbatim}[fontsize=\small, breaklines=true, frame=single]
You are a schedule-generation LLM.
Task: You are given a FIXED, PRE-STRUCTURED multi-day trip plan skeleton ({{DAYS_SKELETON}}). Your ONLY job is to fill empty meal and attraction slots using the provided {{CITIES}} data.

Hard Constraints (STRICT SKELETON ENFORCEMENT):
1. You MUST NOT modify `day`, `current_city`, `transportation`, or `accommodation`.
2. If a field in the skeleton is marked "-", it means that activity is IMPOSSIBLE due to temporal/transport constraints. You MUST keep it as "-".
3. If a field is EMPTY (""), it means the activity IS FEASIBLE. You MUST fill it with a valid entity from the correct city.
4. DO NOT repeat restaurant or attraction names across the entire itinerary.

Entity Selection Logic:
- Meals: Always choose the highest-ranked unused restaurant from `restaurants_ranked` for that specific city.
- Attractions: Select from `attractions_ranked`. 
  - Travel Days: Max 1 attraction.
  - Non-Travel Days: 1-2 attractions (up to 3 if Persona is "Adventure", max 1 if Persona is "Laidback").
- Context: Use {{PERSONA_JSON}} and {{LOCAL_CONSTRAINTS_JSON}} ONLY as soft preferences to guide which valid entities you select. NEVER let preferences override skeleton feasibility.

Formatting Rules:
- Restaurant format: "Restaurant Name, City"
- Attraction format: "Attraction Name, City;" (Multiple attractions separated by space)

Inputs:
- Skeleton: {{DAYS_SKELETON}}
- Ranked Entities: {{CITIES}}
- Persona: {{PERSONA_JSON}} | Constraints: {{LOCAL_CONSTRAINTS_JSON}}
- Original Query: {{query}}

Output strictly in JSON format matching the input skeleton structure:
{
  "days": [
    {
      "day": 1,
      "current_city": "...",
      "transportation": "...",
      "breakfast": "...",
      "lunch": "...",
      "dinner": "...",
      "attraction": "...",
      "accommodation": "..."
    }
  ]
}
\end{Verbatim}

\subsection{POI Scheduler Prompt}
\begin{Verbatim}[fontsize=\small, breaklines=true, frame=single]
You are the POI SCHEDULING AGENT.
Task: Convert a sequence of selected activities (meals, attractions, accommodations) into a strict minute-by-minute timeline for the current travel day.

Authoritative Inputs (DO NOT MODIFY):
- Day Context: {{DAY_TYPE}} (e.g., FIRST_DAY, NON_TRAVEL_DAY, LAST_DAY)
- Fixed Attraction Durations: {{ATTRACTION_DURATIONS_MINUTES}}
- Hard Constraints: BUFFER = 30 mins between activities. MIN_MEAL_GAP = 240 mins.
- Meal Windows: Breakfast (480-630), Lunch (720-940), Dinner (1110-1350).

Execution Mode (Strict Step-by-Step Logic):
You must evaluate each planned activity in sequence using mathematical integer minutes (e.g., 8:00 AM = 480). For each activity:
1. Feasibility Check: Can this activity start at `current_time + BUFFER` and finish within its allowed window (or before the day's transportation cutoff)?
2. Decision: 
   - If feasible: Add POI, update `current_time = POI_END_MIN`, update `last_meal_end` (if meal).
   - If NOT feasible: Skip the activity. Do NOT advance `current_time`.
3. Time Mutation Invariant: Time ONLY advances when an activity is successfully scheduled. Do NOT output "waiting" or "idle" POIs.

Output Format (Strictly Two Sections):

========================
REASONING
========================
- Explain step-by-step which activities were executed or skipped.
- Track `current_time` and `last_meal_end` explicitly.

========================
ITINERARY
========================
<Place Name>, <visit|stay> from <START_MIN> to <END_MIN>;
<Place Name>, <visit|stay> from <START_MIN> to <END_MIN>;

(Note: Output MUST use integer minutes. DO NOT use HH:MM. DO NOT use quotes around names.)
\end{Verbatim}

\subsection{Pros and Cons Extraction Prompt (Review Pipeline)}
\begin{Verbatim}[fontsize=\small, breaklines=true, frame=single]
Instruction: Given 5 reviews for {poi_name}, extract up to 5 Pros and 5 Cons. 
Return EXACTLY this JSON format and nothing else: {"Pros": [], "Cons": []}
Do not use markdown blocks. Be concise.

Reviews:
{reviews_text}

Output:
\end{Verbatim}

\subsection{Skeleton Validation / Repair Prompt}
\begin{Verbatim}[fontsize=\small, breaklines=true, frame=single]
You previously generated an INVALID schedule. 
Fix ONLY the listed issues. Do NOT change transportation, current_city, or accommodation.

=== SKELETON ===
{skeleton_json}

=== YOUR PREVIOUS OUTPUT ===
{previous_output_json}

=== VALIDATION ERRORS ===
{validation_errors_list}
(Example format: "1. Day 3 – attraction | Reason: Attraction required by skeleton | Allowed options: [...]")

INSTRUCTIONS:
- Fix ONLY the fields mentioned in the validation errors.
- For any field NOT mentioned, you MUST copy the value EXACTLY from your PREVIOUS OUTPUT.
- If allowed options are provided, use ONLY those options.
- If the skeleton value is "", you MUST fill it with a valid entity from the correct city.
- If the skeleton value is "-", you MUST keep it as "-".
- Do NOT repeat attractions and restaurant names across the itinerary.
- Do NOT add, remove, or rename JSON keys.

Output strictly in JSON format:
{
  "days": [ ... corrected day objects ... ]
}
\end{Verbatim}

\subsection{LLM-as-a-Judge Evaluation Prompt}
\label{app:llm_judge_prompt}
\begin{Verbatim}[fontsize=\small, breaklines=true, frame=single]
You are an expert evaluator for personalized travel planning.

The two itineraries are presented in randomized order.

Do not assume one itinerary is better simply because:
- it contains more entities,
- includes more review evidence,
- or contains more detailed descriptions.

Focus strictly on experiential alignment with the traveler persona.

Your task is to compare two itineraries generated for the SAME traveler query.

Focus specifically on:
- budget alignment
- travel style alignment
- ambiance and comfort
- cultural exploration quality
- avoidance of poor user experiences

===========================================
TRAVELER PERSONA
===========================================

{persona}

===========================================
ITINERARY A
===========================================

{itinerary_a}

===========================================
REVIEW EVIDENCE FOR ITINERARY A
===========================================

{reviews_a}

===========================================
ITINERARY B
===========================================

{itinerary_b}

===========================================
REVIEW EVIDENCE FOR ITINERARY B
===========================================

{reviews_b}

===========================================
EVALUATION RUBRIC
===========================================

Score both itineraries from 1-10 on:

1. Persona Alignment
2. Experiential Quality
3. Risk Avoidance
4. Overall Satisfaction

Then provide a pairwise preference decision:

- A strongly preferred
- A preferred
- Tie
- B preferred
- B strongly preferred

===========================================
OUTPUT FORMAT
===========================================

Return STRICT JSON format.
Do not include markdown formatting.
Return valid JSON only.

{
    "persona_alignment": {
        "A": score,
        "B": score
    },
    "experiential_quality": {
        "A": score,
        "B": score
    },
    "risk_avoidance": {
        "A": score,
        "B": score
    },
    "overall_satisfaction": {
        "A": score,
        "B": score
    },
    "pairwise_preference": "...",
    "reasoning": "short explanation"
}
\end{Verbatim}

\end{document}